\documentclass[twoside,11pt]{article}

\usepackage{amsthm}

\usepackage{jmlr2e}

\usepackage{enumerate}
\usepackage{comment}
\usepackage{physics}
\usepackage{mathtools}
\usepackage[utf8]{inputenc} 
\usepackage[T1]{fontenc}    
\usepackage{hyperref}       
\usepackage{url}            
\usepackage{booktabs}       
\usepackage{amsfonts}       
\usepackage{nicefrac}       
\usepackage{microtype}      
\usepackage{wrapfig}

\DeclarePairedDelimiter{\ceil}{\lceil}{\rceil}
\DeclarePairedDelimiter{\floor}{\lfloor}{\rfloor}

\DeclareMathOperator{\Var}{Var}

\DeclareMathOperator{\supp}{Supp}

\DeclareMathOperator{\argmin}{argmin}

\DeclareMathOperator{\card}{card}
\DeclareMathOperator{\mul}{mul}
\DeclareMathOperator{\pol}{pol}
\DeclareMathOperator{\cut}{cut}

\DeclareMathOperator{\tlr}{tlr}

\newcommand{\1}{\mbox{1}\hspace{-0.25em}\mbox{l}}
\newcommand{\defeq}{\coloneqq}
\newcommand{\eqdef}{\eqqcolon}
\newcommand{\R}{\mathbb{R}}

\newcommand{\N}{\mathbb{N}}

\newcommand{\mB}{\mathcal{B}}
\newcommand{\mC}{\mathcal{C}}

\newcommand{\mF}{\mathcal{F}}
\newcommand{\mG}{\mathcal{G}}
\newcommand{\mH}{\mathcal{H}}
\newcommand{\mI}{\mathcal{I}}

\newcommand{\mM}{\mathcal{M}}
\newcommand{\mN}{\mathcal{N}}
\newcommand{\mO}{\mathcal{O}}
\newcommand{\mP}{\mathcal{P}}
\newcommand{\mS}{\mathcal{S}}

\newcommand{\mX}{\mathcal{X}}

\newcommand{\I}{\mathrm{I}}

\newcommand{\dm}{\dd{\mu}}

\newcommand{\regularity}{\dim_{\textup{R}}}
\newcommand{\minkowski}{\dim_{\textup{M}}}

\newcommand{\lminkowski}{\underline{\dim}_{\textup{M}}}

\renewcommand{\tilde}{\widetilde}
\renewcommand{\hat}{\widehat}

\allowdisplaybreaks

\usepackage{color}

\newcommand{\bc}{\color{black}}

\usepackage{lastpage}
\jmlrheading{21}{2020}{1-\pageref{LastPage}}{1/20; Revised
7/20}{9/20}{20-002}{Ryumei Nakada and Masaaki Imaizumi}
\ShortHeadings{Adaptivity of Deep Neural Network with Intrinsic Dimensionality}{Nakada and Imaizumi}

\begin{document}

\title{Adaptive Approximation and Generalization of\\Deep Neural Network with Intrinsic Dimensionality}

\author{\name Ryumei Nakada \email ryumei.nakada@rutgers.edu \\
       \addr Department of Statistics, Rutgers University, USA\\
       The University of Tokyo, Japan
       \AND
       \name Masaaki Imaizumi \email imaizumi@g.ecc.u-tokyo.ac.jp \\
       \addr Komaba Institute for Science, The University of Tokyo, Japan\\
       Center for Advanced Intelligence Project, RIKEN, Japan}

\editor{Rina Foygel Barber}

\maketitle

\begin{abstract}
In this study, we prove that an intrinsic low dimensionality of covariates is the main factor that determines the performance of deep neural networks (DNNs). DNNs generally provide outstanding empirical performance. Hence, numerous studies have actively investigated the theoretical properties of DNNs to understand their underlying mechanisms. In particular, the behavior of DNNs in terms of high-dimensional data is one of the most critical questions. However, this issue has not been sufficiently investigated from the aspect of covariates, although high-dimensional data have practically low intrinsic dimensionality. In this study, we derive bounds for an approximation error and a generalization error regarding DNNs with intrinsically low dimensional covariates. We apply the notion of the Minkowski dimension and develop a novel proof technique. Consequently, we show that convergence rates of the errors by DNNs do not depend on the nominal high dimensionality of data, but on its lower intrinsic dimension. We further prove that the rate is optimal in the minimax sense. We identify an advantage of DNNs by showing that DNNs can handle a broader class of intrinsic low dimensional data than other adaptive estimators. Finally, we conduct a numerical simulation to validate the theoretical results.
\end{abstract}


\section{Introduction}

\textit{Deep neural networks} (DNNs) \citep{lecun2015deep,goodfellow2016deep} have attracted considerable attention as statistical modeling for deep learning, owing to favorable outcomes of deep learning in various applications \citep{collobert2008unified,he2016deep}.
We often observe that prediction and estimation by DNNs can achieve higher accuracy than that by several standard conventional methods, such as kernel methods \citep{schmidhuber2015deep}. 
To understand the performance of DNNs and effectively exploit them, numerous studies have extensively investigated their theoretical aspects, such as the approximation power of DNNs \citep{yarotsky2017error,arora2018stronger,bartlett2017spectrally,schmidt2017nonparametric}.

A nonparametric regression problem is one of the standard frameworks for investigating the mechanisms of DNNs  \citep{bauer2019deep,schmidt2017nonparametric,kohler2019deep,imaizumi2018deep,suzuki2018adaptivity,schmidt2019deep}.
Suppose we have a set of $n$ observations $\{(X_i, Y_i)\}_{i=1}^n \subset [0,1]^D \times \R$ which is independently and identically generated from the regression model
\begin{align}
  Y_i = f_0(X_i) + \xi_i, ~X_i \sim \mu, ~~i=1,...,n, \label{eq:model}
\end{align}
where the covariates $X_i$ marginally follow a probability measure $\mu$, and $\xi_i$ is an independent and identically distributed (i.i.d.) Gaussian noise that is independent of $X_i$, $E[\xi_i] = 0$ and $E[\xi_i^2] = \sigma^2$, where $\sigma > 0$.
The aim of this study is to estimate the unknown function $f_0:[0,1]^D \to \R$ by an estimator $\hat{f}$ applied to DNNs.
To measure a performance of DNNs, we consider the following value:
\begin{align}
    \|\hat{f} - f_0\|_{L^2(\mu)}^2 = E_{X \sim \mu}[(\hat{f}(X) - f_0(X))^2], \label{def:gen}
\end{align}
with the marginal measure $\mu$.
The error is also known as a generalization error and is generally used to evaluate the performance of DNNs.

\textit{The curse of dimensionality} is one of the most significant problems with the DNN framework, in which the theoretical performance of DNNs deteriorates as data dimensionality increases.
For the regression problem, the generalization error of DNNs is on the order $\tilde{O}( n^{-2\beta / (2\beta + D)} )$, where $\beta > 0$ is the degree of smoothness of $f_0$.
The rate is known to be optimal in the minimax sense in a typical setting  \citep{schmidt2017nonparametric}.
Since $D$ tends to be very large in machine learning applications (e.g., the number of pixels in images), the theoretical generalization error decreases very slowly as $n$ increases.
Thus, the theoretical bound for the error is quite loose for describing the empirical performance of DNNs.
To avoid the slow convergence rate, numerous studies have investigated specific structures of functions as a target of approximation when using DNNs. 
One typical approach is to consider various types of smoothness or spectral distribution; many general notions of smoothness are investigated \citep{barron1993universal,barron1994approximation,barron2018approximation,montanelli2017deep,suzuki2018adaptivity,schmidt2017nonparametric}.
Alternatively, it is also common to introduce a specific form of $f_0$.
For example, when $f_0$ takes on the form of conventional statistical models (e.g., the generalized single index or multivariate adaptive regression splines (MARS)), the convergence rates of DNNs evidently improve \citep{bauer2019deep,kohler2019deep}.
Additionally, when $f_0$ has a functional form involving manifolds, the convergence rate of DNNs depends on the manifold dimensions \citep{schmidt2019deep}.


In contrast to the studies focused on $f_0$, the behavior of the measure $\mu$ of covariates has not been well studied despite several significant motivations for the investigation of   $\mu$ exist.
First, we frequently observe that high-dimensional data have an implicit structure such as lying around low dimensional sets (e.g., manifolds) in practice \citep{tenenbaum2000global,belkin2003laplacian}.
We numerically confirm that several well-known real data have approximately $30$ intrinsic dimensions, while their nominal dimensions are approximately $1,000$ (see Section \ref{section:empirical}).
Since the low intrinsic dimensionality is well known in the field of machine learning, several well-designed methodologies use this empirical fact \citep{masci2015geodesic,arjovsky2017towards}.
Second, the literature on nonlinear dimension reduction indicates that the low intrinsic dimensionality of covariates can be a crucial factor in overcoming the curse of dimensionality.
Several estimators, such as kernel methods and the Gaussian process regression, have been shown to achieve a fast convergence rate depending only on their intrinsic dimensionality \citep{bickel2007local,kpotufe2011k,kpotufe2013adaptivity,yang2016bayesian}.
Despite these motivations, connecting DNNs with an intrinsic dimension of data is a nontrivial task; thus, investigating this property remains an important open question.

In this study, we investigate the performance of DNNs regarding $D$-dimensional data, which have a $d$-dimensional intrinsic structure such that $d < D$.
To describe the intrinsic dimensionality of data, we apply the notion of \textit{Minkowski dimension}.
Moreover, we develop a proof technique to adapt DNNs to the intrinsic low dimensional structure.
Consequently, we derive the rates of the approximation and generalization errors in DNNs, which depend only on $d$ and $\beta$, but not on $D$, as summarized in Table \ref{tab:rates}.
In summary, we prove that the convergence speed of DNNs is independent of the nominal dimension of data, but does on their intrinsic dimension.
We also prove that the derived rate is optimal in the minimax sense and, finally, verify the theoretical results using numerical experiments.

\begin{table}[htbp]
\begin{center}
\begin{small}
\begin{sc}
\begin{tabular}{c|cc}
\toprule
 & Approximation error & Generalization error \\
\midrule
Existing    & $\mO(W^{-\beta/D})$ & $\tilde{\mO}(n^{-2\beta/(2\beta + D)})$ \\
\textbf{Ours} ($d$-Minkowski dim.) & ${\mO}(W^{-\beta/d})$  & $\tilde{\mO}(n^{-2\beta/(2\beta + d)})$  \\
\bottomrule
\end{tabular}
\end{sc}
\end{small}
\end{center}
\caption{Derived rates of approximation and generalization errors by DNNs with $W$ parameters and $n$ observations.
$\beta > 0$ denotes the smoothness of the generating function,  and $D$ is dimension of the observations, and $d$ is an intrinsic dimension of $\mu$.
In general, $d \leq D$ holds. \label{tab:rates}}
\end{table}

Our results describe an advantage inherent in DNNs compared to several other methods.
That is, we demonstrate that DNNs can achieve a fast convergence rate over a broader class of data distributions.
Some adaptive methods, such as the kernel and Gaussian process estimators, can achieve a fast convergence rate with the intrinsic dimension.
However, they can achieve a fast rate only when data are on smooth manifolds or are generated from doubling measures.
In contrast, we show that DNNs can achieve a fast rate over a broader class of $\mu$, such as data on highly non-smooth fractal sets.
This advantage is due to the use of the Minkowski dimension, which can cover a broader class of distributions.

As a technical contribution of this study, we develop a proof for DNNs with an optimal partition of hypercubes in the domain to handle the Minkowski dimension.
To evaluate an error regarding the dimension, we have to follow the two steps: (i) divide the domain of $f^0$ into hypercubes and (ii) combine sub-neural networks in each of the hypercubes.
However, a naive combination makes the depth (number of layers) of DNNs diverge because of the accumulation of errors of adjacent hypercubes.
In our proof, to avoid this problem, we develop a particular set of partitions for a set of hypercubes and then unify the sub-neural networks within each of the partitions.
Using this technique, we can avoid the accumulation of errors to achieve the desired convergence rate.

We summarize the contributions of this study as follows:
\begin{enumerate}
\setlength{\parskip}{0cm}
  \setlength{\itemsep}{0cm}
    \item We prove that DNNs can avoid the curse of dimensionality by adapting to the intrinsic low dimensionality of the data with the Minkowski dimension.
    \item We present a relative advantage of DNNs, in which a fast convergence rate with a broader class of distributions can be achieved compared to other methods that are also adaptive to an intrinsic dimension.
    \item As proof, we derive rates of approximation and generalization errors for $D$-dimensional data, which have $d$ intrinsic dimension, and demonstrate that the rate is minimax optimal.
\end{enumerate}

The remainder of this study is organized as follows.
Section \ref{sec:low-dim} discusses the notion of intrinsic dimensionality and defines the Minkowski dimension.
Section \ref{sec:approx} shows an upper bound for the approximation error of DNNs.
Section \ref{sec:gen} provides the upper and lower bounds of the generalization error of DNNs.
Section \ref{sec:compare} compares our main results with several related studies involving DNNs and other methods.
Section \ref{sec:numerical} provides experimental evidence to support the theoretical results.
Finally, Section \ref{sec:conclusion} summarizes our conclusions.
The appendix includes a full version of the proof.

\subsection{Basic Notation}

For an integer $z$, $[z]:=\{1,2,...,z\}$ is a set of positive integers no greater than $z$.
For a vector $b \in \R^d$, $\norm{b}_q := (\sum_{j=1,...,d} b_j^q)^{1/q}$ is a $q$-norm for $q \in [0,\infty]$.
For a measure $\mu$, the support of $\mu$ is written as $\supp(\mu)$.
For a function $g: \R^D \to \R$, $\norm{g}_{L^p(\mu)} \defeq (\int g^p \dm)^{1/p}$ is the $L^p(\mu)$ norm, with a probability measure $\mu$.
$\tilde{O}(\cdot)$ is the Landau's big O, ignoring a logarithmic factor.
With $\varepsilon > 0$, $\mN(\Omega, \varepsilon)$ is the fewest number of $\varepsilon$-balls that cover $\Omega$ in terms of $\|\cdot\|_{\infty}$.
For a measure $\mu$, $\supp(\mu)$ denotes the support for $\mu$.

\section{Intrinsic Low Dimensionality of Covariates} \label{sec:low-dim}

\subsection{Empirical Motivation} \label{section:empirical}

As the motivation for considering low intrinsic dimensionality, we provide an empirical analysis of real datasets, such as handwritten letter images using the modified National Institute of Standards and Technology (MNIST) dataset \citep{lecun2015deep} and object images using the Canadian Institute for Advanced Research (v) dataset  \citep{krizhevsky2009learning}.
Since the data are images, their nominal dimension $D$ is equal to the number of pixels in each image.
We apply several dimension estimators, such as the local principal component analysis (LPCA) \citep{fukunaga1971algorithm,bruske1998intrinsic}, the method with maximum likelihood method (ML) \citep{haro2008translated}, and the expected simplex skewness (ESS) \citep{johnsson2015low} to estimate the intrinsic dimensions of $30,000$ samples from each of the datasets.
The results in Table \ref{fig:dim} indicate that the estimated intrinsic dimensions are significantly less than $D$.
Although the definitions of intrinsic dimensions are not standard, the results provide motivation to conduct an in-depth investigation on low intrinsic dimensionality.

\begin{table}[h]
\begin{center}
\begin{small}
\begin{sc}
\begin{tabular}{l|c|ccc}
\toprule
&&\multicolumn{3}{c}{Intrinsic Dimension $d$}\\
Data set & $D$ & LPCA & ML & ESS \\
\midrule
MNIST    & 784 & 37 & 13.12 & 29.41\\
CIFAR-10 & 1024 & 9 & 25.84 & 27.99\\
\bottomrule
\end{tabular}
\end{sc}
\end{small}
\end{center}
\caption{Estimated intrinsic dimensions of the MNIST and CIFAR-10 datasets. The dimensions are estimated from $30,000$ sub-samples from the original datasets. \label{fig:dim}}
\end{table}

\subsection{Notion of Intrinsic Dimensionality}

\begin{figure}[htbp]
        \centering
        \includegraphics[width=0.99\hsize]{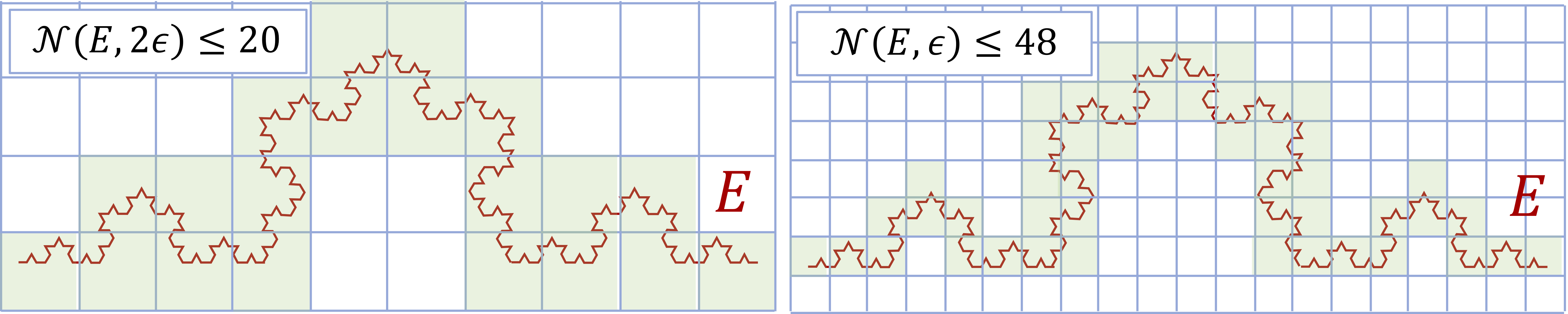}
        \caption{The Koch curve $E$ (red lines) and covering $\max$-balls (green squares) for $E$.
        The Minkowski dimension of $E$ is $d = \log 4 / \log 3 \approx 1.26$, while $E$ is a subset of $\R^D$ with $D=2$.
        \label{fig:mink_dim}}
\end{figure}

We introduce the notion of dimensionality in this study.
Although there are numerous definitions for dimensionality, we employ the following general notion.
\begin{definition}[Minkowski Dimension]
  The (upper) Minkowski dimension of a set $E \subset [0,1]^D$ is defined as
  \begin{align*}
       \minkowski E \defeq \inf\bigl\{d^* \geq 0 \mid \limsup_{\varepsilon \downarrow 0} \mN(E, \varepsilon) \varepsilon^{d^*} = 0\bigr\}.
  \end{align*}
\end{definition}
The Minkowski dimension measures how the number of covering balls for $E$ is affected by the radius of the balls.
Since the dimension does not depend on smoothness, it can measure the dimensionality of highly non-smooth sets, such as fractal sets (e.g., Koch curve).
Figure \ref{fig:mink_dim} shows an image of how the covering balls measure the Minkowski dimension of $E$.

\textbf{Relationship to Other Dimensions:} The Minkowski dimension can describe a broader class of low dimensional sets compared to several other dimensionalities.
For example, the notion of \textit{manifold dimension} describes the dimensionality of sets with smooth structures and is one of the most common notions used to describe an intrinsic dimensionality \citep{bickel2007local,yang2016bayesian}.
While manifold dimensions are valid only for smooth sets such as circles, we can apply the Minkowski dimension to sets without such restriction.
Consequently, for a set $E \subset [0,1]^D$, we show that
\begin{align*}
    \{E  \mid \dim_M E \leq d\} \supset \{E  \mid E\mbox{~is~a~$d$-dimensional~manifold}\},
\end{align*}
holds (see Lemma \ref{lem:manifolds} in Section \ref{sec:dimensions}).
In addition, the notion of \textit{regularity dimensions} is used for an intrinsic dimensionality \citep{kpotufe2011k,kpotufe2013adaptivity}.
Similar to the manifold case, the Minkowski dimension is a more general notion than the regularity dimension (see Lemma \ref{lem:mink_regular} in Section \ref{sec:dimensions}).

\section{Approximation Results} \label{sec:approx}

\subsection{Preparation}

To investigate an approximation power of DNNs, we provide a rigorous formulation.
Here, we consider DNNs with a rectified linear unit (ReLU) activation function $\rho(x_1, \dots, x_p) := (\max\qty{x_1, 0}, \dots, \max\qty{x_p, 0})$.

\begin{definition}[Deep Neural Networks]
  Let $L$ be a number of layers. For each $\ell \in [L]\cup \{0\}$, $p_\ell \in \N$ be a number of nodes for each layer $\ell$, and $A_{\ell} \in \R^{p_{\ell} \times p_{\ell-1}}$ and $b_\ell \in \R^{p_\ell}$ be a parameter matrix and vector.
  Let $\rho_{b_\ell}:= \rho (\cdot + b_\ell)$ be a shifted ReLU activation.
  Then, the realization of a neural network architecture $\Phi := ((A_{L}, b_{L}), \dots, (A_1, b_1))$ is denoted as $R(\Phi) : \R^{p_0} \to \R^{p_{L}}$, which has a form
\begin{align}
    R(\Phi)(x) = A_{L} \rho_{b_{L-1}}\circ \cdots \circ A_{2} \rho_{b_{1}} (A_{1}x) + b_{L},\mbox{~for~}x \in [0,1]^{p_0}. \label{def:dnn}
\end{align}
$R(\Phi)(x)_i$ denotes the $i$-th output of $R(\Phi)(x)$.
\end{definition}

Further, we define a set of realizations of DNNs with several restrictions.
Specifically, we bound a number of layers, parameters, and its scale.
For each $\Phi$, the number of layers of $\Phi$ is written as $L(\Phi)$, a number of parameters of $\Phi$ is $W(\Phi) \defeq \sum_{\ell=1}^L \|b_\ell\|_0 +   \|\mbox{vec}(A_\ell)\|_0$, and the scale of parameters of $\Phi$ is $B(\Phi) = \max_{\ell = 1,...,L} \max\{\|b_\ell\|_{\infty} ,  \|\mbox{vec}(A_\ell)\|_{\infty}\}$.

\begin{definition}[Functional Set by DNNs]
  With a tuple $(W', L',B')$, a functional set by DNNs is defined as follows:
\begin{align*}
    &\mF(W', L', B') = \bigl\{R(\Phi): [0,1]^{p_0} \to \R^{p_L}  \mid L(\Phi) \leq L', W(\Phi) \leq W', B(\Phi) \leq B'\bigr\}.
\end{align*}
\end{definition}

To discuss the approximation power of DNNs, we define the \textit{H\"older space} as a family of smooth functions.
For a function $f: \R^D \to \R$, $\partial_d f(x)$ is a partial derivative with respect to a $d$-th component, and $\partial^\alpha f := \partial_1^{\alpha_1}\cdots \partial_D^{\alpha_D}f$ using multi-index $\alpha = (\alpha_1,...,\alpha_D)$.
For $z \in \R$, $\lfloor z \rfloor$ denotes the largest integer that is less than $z$.

\begin{definition}[H\"older space]
Let $\beta > 0$ be a degree of smoothness.
For $f:[0,1]^D \to \R$, the \textit{H\"older norm} is defined as
\begin{align*}
   &\norm{f}_{\mH(\beta, [0, 1]^D)} :=\max_{\alpha: \norm{\alpha}_1 < \floor{\beta}} \sup_{x \in [0,1]^D}|\partial^\alpha f(x)|  +  \max_{\alpha: \norm{\alpha}_1 = \floor{\beta}} \sup_{x, x' \in [0, 1]^D, x\neq x'} \frac{ \abs{ \partial^\alpha f(x) - \partial^{\alpha} f(x')} }{\norm{x - x'}_{\infty}^{ \beta - \floor{\beta} } } .
\end{align*}
Then, the \textit{H\"older space} on $[0,1]^D$ is defined as
\begin{align*}
    \mH(\beta, [0, 1]^D) = \qty{f \in C^{\floor{\beta}}([0, 1]^D) \middle\vert \norm{f}_{\mH(\beta, [0, 1]^D)} < \infty}.
\end{align*}
Also, $\mH(\beta, [0, 1]^D, M) = \qty{f \in \mH(\beta, [0, 1]^D) \middle\vert \norm{f}_{\mH(\beta, [0, 1]^D)} \leq M}$ denotes the $M$-radius closed ball in $\mH(\beta, [0, 1]^D)$.
\end{definition}

\subsection{Approximation with Low Dimensionality}

We evaluate how well DNNs approximate a function $f_0 \in \mH(\beta, [0, 1]^D)$ with a probability measure $\mu$ whose support has a Minkowski dimension less than $d$.
That is, we measure an approximation error using the norm $\norm{\cdot}_{L^\infty(\mu)}$ with $\mu$ as a base measure.

\begin{theorem}[Approximation with Minkowski dimension] \label{thm:approx_mink}
  Suppose $d > \minkowski \supp(\mu)$ holds with $d < D$.
  {\bc
  For $\beta,M > 0$, we define $r := 2 + \floor{(1 + \floor{\beta})/(2d)}$, $c_{\beta, D, d} := 384 \beta (11+(1+\beta)/d) ( 36 r + 83 + 6 \cdot 4^{r+2})\cdot 6^{d/(1+\floor{\beta})}$, $c_\mu := \sup_{\varepsilon > 0} \mN(\supp\mu, \varepsilon) \varepsilon^d$, and constants $C_1, C_2 > 0$ such as
  \begin{align*}
      &C_1\leq \qty(4c_\mu(50D + 17 + 8D^{2 + \floor{\beta}} c_{\beta, D, d})  D^d (3 M)^{d/\beta} + D^{2 + \floor{\beta}} c_{\beta, D, d} 2^{d/\beta+5}),\\
      &C_2 \leq 11 + 6D  + (11+(1+\beta)/d) (1 + (1 \vee \ceil{\log_2 \beta})).
  \end{align*}
  Also, for $\varepsilon > 0$, we consider a triple $(W, L, B)$ with
  \begin{align}\label{eq:triple}
      &W = C_1 \varepsilon^{-d/\beta}, ~ L = C_2, \mbox{~and~}B = O( \varepsilon^{-s}).
  \end{align}
  }
  Then, for sufficiently small $\varepsilon_0 > 0$, any $\varepsilon \in (0, \varepsilon_0)$ and any $f_0 \in \mH(\beta, D, M)$, we obtain
  \begin{align*}
      \inf_{R(\Psi) \in \mF(W, L, B)}\|R(\Psi) - f_0\|_{L^\infty(\mu)} \leq \varepsilon.
  \end{align*}
\end{theorem}

The following corollary summarizes the result in Theorem \ref{thm:approx_mink}.
\begin{corollary}[Approximation Rate]
    With the triple $(W,L,B)$ as defined in Theorem \ref{thm:approx_mink}, an existing $R(\Psi) \in \mF(W,L,B)$ satisfies for sufficiently large $W$,
    \begin{align*}
        \|R(\Psi) - f_0\|_{L^\infty(\mu)} = \mO(W^{-\beta/d}).
    \end{align*}
\end{corollary}

The result indicates that the order $\mO(W^{-\beta/d})$ depends only on $d$ and $\beta$, but not on $D$.
That is, the approximation rate behaves as if the data are $d$-dimensional, although they are nominally $D$-dimensional.

{\bc

Additionally, the results of Theorem 2 suggest the following intuitions.
First, a finite number of layers is sufficient to achieve the convergence rate, because $L$ does not diverge with small $\varepsilon$.
Second, the constant terms $C_1$ and $C_2$ in Theorem \ref{thm:approx_mink} depend on $D$ polynomially.
To achieve the results, we develop an additional proof technique to achieve it with for low intrinsic dimensionality.
}

\vspace{\baselineskip}

\begin{figure}[htbp]
    \centering
    \includegraphics[width=0.45\hsize]{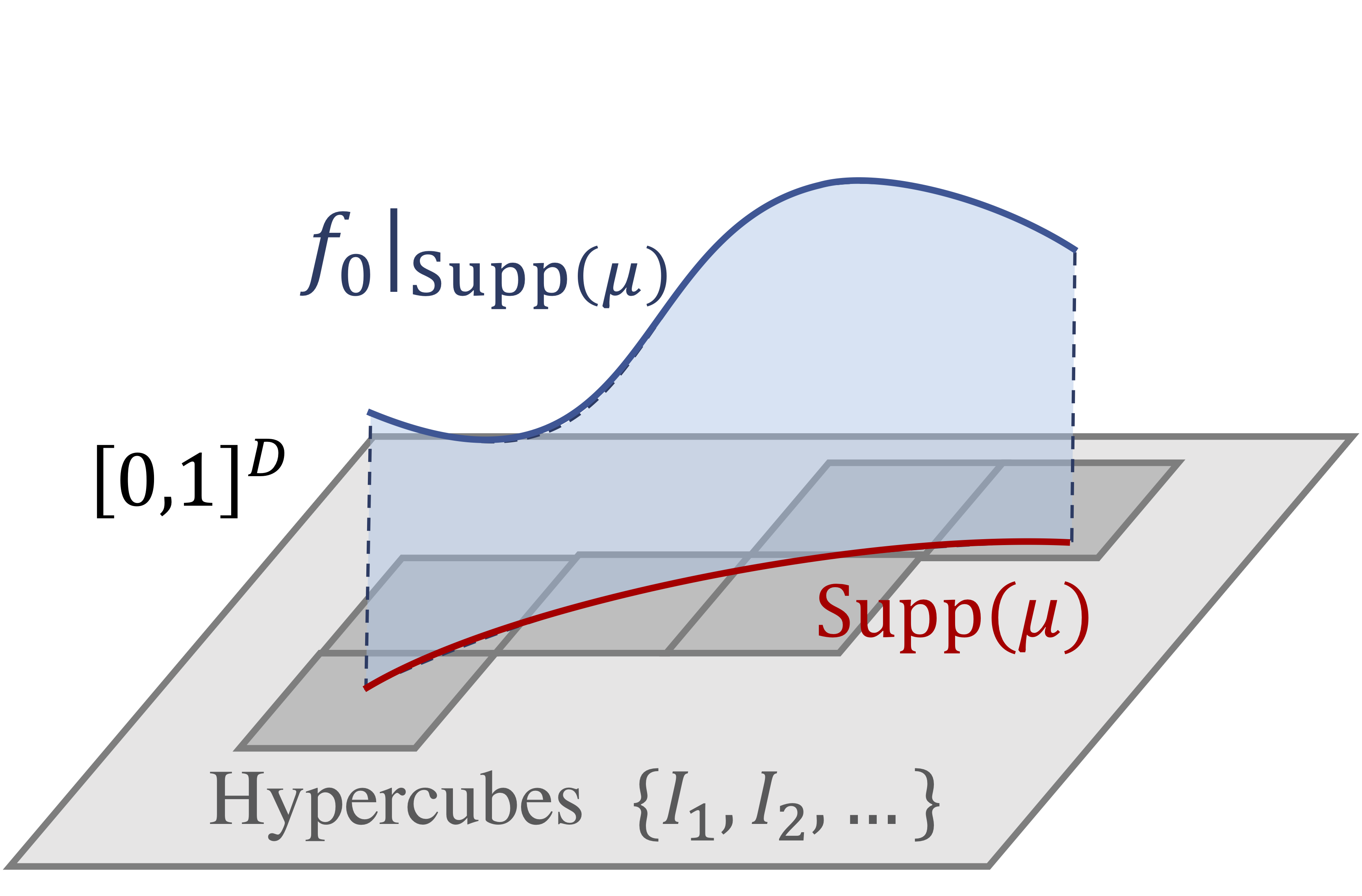}
    \includegraphics[width=0.45\hsize]{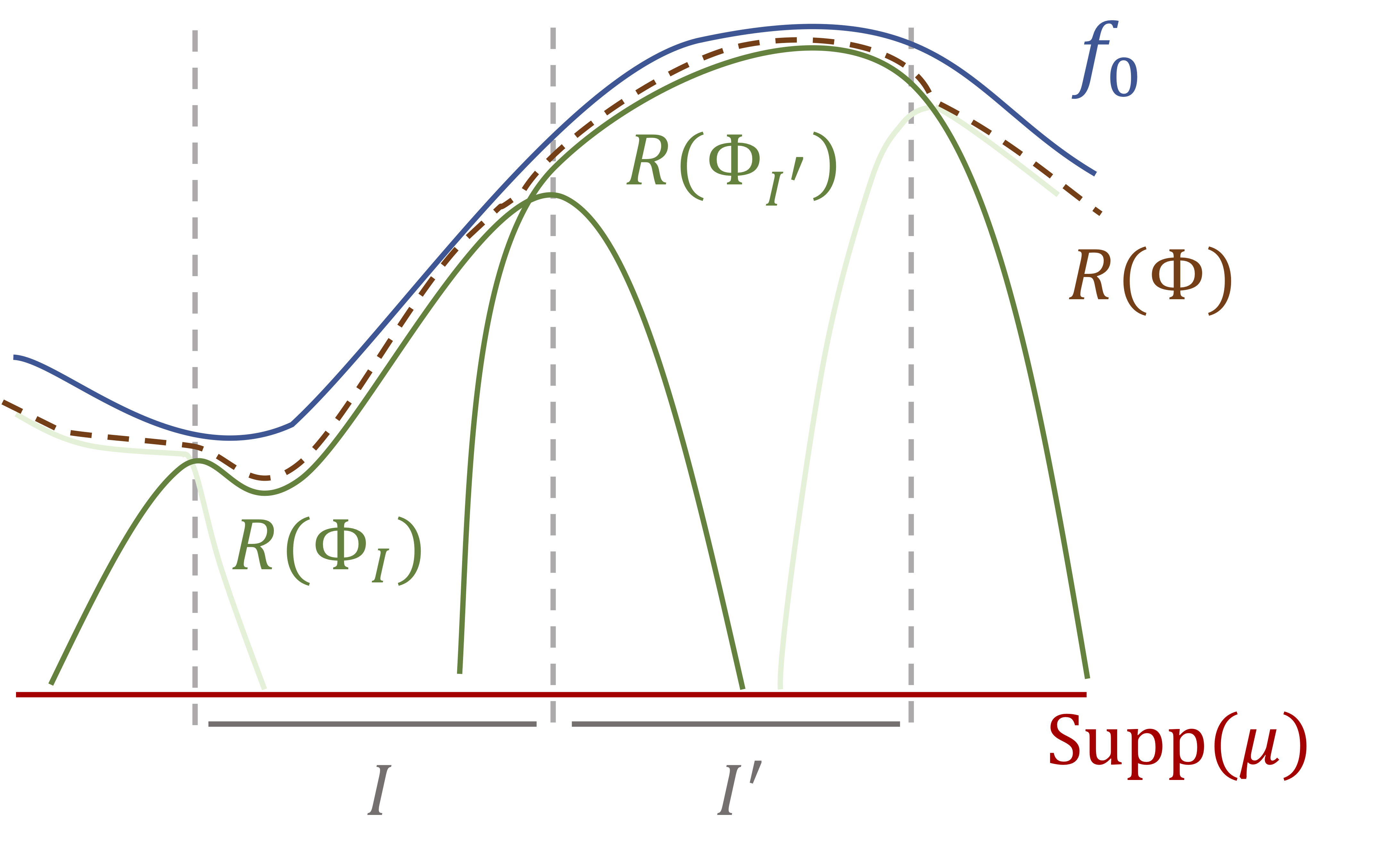}
    \caption{An illustration of our proof.
    [Left] $\supp(\mu)$ (the red curve) in $[0,1]^D$, where $f_0$ (the blue curve) is restricted to $\supp(\mu)$.
    The hypercubes (the gray squares) $\mI=\{I_1,I_2,...\}$ cover $\supp(\mu)$.
    DNNs approximate $f_0$ within each of the hypercubes.
    [Right] Within hypercubes $I,I' \in \mI$, the DNNs produce a function that approximates $f_0$ locally.
    Using the trapezoid-type approximators $R(\Phi_I)$ and $R(\Phi_{I'})$ (the green curves), we define the approximator $R(\Phi)$ (the brown curve) as the maximum of the trapezoid approximators.}
    \label{fig:proof_approx}
\end{figure}

\noindent \textbf{Proof Outline of Theorem \ref{thm:approx_mink}}:
Let $\mI$ be a minimum set of hypercubes of side length $\gamma$ covering $\supp(\mu)$.
We partition $\mI$ into $\mI_1, \dots, \mI_K$ such that each subset $\mI_k$ consists of hypercubes separated by $\gamma$ from each other.
For $I \in \mI$, let $R(\Phi_I)$ be a trapezoid-type approximator represented by a neural network $\Phi_I$ that approximates $f_I \1_I$ (the green curve in the right panel in Figure \ref{fig:proof_approx}), where $f_I$ is the Taylor expansion of $f_0$ around any point in $I$. We note that for any $\mI_k$, support of $R(\Phi_I)$ for $I \in \mI_k$ are disjoint.
Then, we define a neural network $\Phi$ to realize $R(\Phi) = \max_{1\leq k \leq K} \sum_{I' \in \mI_k} R(\Phi_{I'})$ as our novel approximator.
For $x \in I$ for some $I \in \mI$, we have
\begin{align*}
    \abs{R(\Phi)(x) - f_0(x)} \leq \underbrace{\max_{I' \in \Xi(I)} \abs{R(\Phi_{I'})(x) - f_{I'}(x)}}_{T_1} + \underbrace{\max_{I' \in \Xi(I)} \abs{f_{I'}(x) - f_0(x)}}_{T_2}
\end{align*}
where $\Xi(I)$ denotes the set of hypercubes neighbouring $I$, including $I$ itself.
The inequality holds because $R(\Phi_{I'})(x) = 0$ holds for all $I' \not \in \Xi(I)$ and thus $\sum_{I' \in \mI_k} R(\Phi_{I'})(x) = R(\Phi_{I''})(x)$ for some $I'' \in \mI_k \cap \Xi(I)$.
To bound the term $T_1$, we evaluate the trapezoid approximation using the \textit{sawtooth} approximation \citep{telgarsky2015representation} for the Taylor polynomials.
$T_2$ is evaluated by the Taylor approximation for $\mH(\beta, [0, 1]^D, M)$.
Regarding the effect of the partition $\mI_k$, we limit $K$ by a constant depending only on $D$.
Hence, we show that the number of layers can be finite.
Since $\minkowski\supp(\mu) \leq d$ holds, there are $O(\gamma^{-d})$ hypercubes used to approximate $f_0$ on $\supp(\mu)$.
An image of the entire procedure is provided in Figure \ref{fig:proof_approx}.

\section{Generalization Results} \label{sec:gen}

We investigate the generalization error of DNNs using a nonparametric regression problem.
Suppose we have a set of $n$ observations $\{(X_i, Y_i)\}_{i=1}^n$ from the regression model \eqref{eq:model} with $f_0 \in \mH(\beta, [0, 1]^D, M)$, the marginal measure $\mu$,  and Gaussian noise $\xi_i$.
From the observations, we introduce an estimator for $f_0$.
The estimator $\hat f$ is defined as $\hat f(x) := \max\{-C_B, \min\{ C_B, \tilde f(x)\}\}$, where
\begin{align}
    \tilde{f} \in \argmin_{f \in \mF(W, L, B)} \sum_{i=1}^n ( Y_i - f(X_i))^2. \label{def:fhat}
\end{align}
$C_B > 0$ denotes a threshold for the clipped estimator $\tilde f$.
We note that calculating $\hat{f}$ is not straightforward because the loss function in \eqref{def:fhat} is non-convex.
However, we employ the estimator $\hat{f}$, because we aim to investigate the generalization error in terms of $n$, which is independent of the difficulty of optimization.
We can obtain an approximated version of $\hat{f}$ using various optimization techniques, such as multiple initializations or Bayesian optimization.

\subsection{Generalization Error with Low Dimensionality}

We provide a generalization error of $\hat{f}$ for the case when $\supp(\mu)$ has a low Minkowski dimension.

\begin{theorem}[Generalization with Minkowski dimension] \label{thm:gen_mink}
  Fix any $f_0 \in \mH(\beta, [0, 1]^D, M)$ and suppose $d > \minkowski\supp(\mu)$.
  Set a triple $(W,L,B)$ with the constants $C_1, C_2$ and $s$ appearing in Theorem \ref{thm:approx_mink} as $W = C_1 n^{d/(2\beta + d)}$, $L = C_2$, and $B = O(n^{2\beta s/(2\beta + d)}\log n)$.
  Then, there exists a constant $C = C(c_\mu, \beta, D, d, M, \sigma)$ such that
  \begin{align*}
    &\|\hat f - f_0\|_{L^2(\mu)}^2 \leq C n^{-2\beta/(2\beta + d)}(1 + \log n)^2
  \end{align*}
  holds with probability at least $1 - 2\exp(-n^{d/(2\beta + d)})$ for any $n \geq N$ and $C_B \geq \|f_0\|_{L^\infty(\mu)}$ with a sufficiently large $N$.
\end{theorem}

The derived generalization error is on the order of $\tilde{O}(n^{-2\beta / (2\beta + d)})$, which is independent from $D$.
That is, we show that the convergence rate of DNNs is determined by the intrinsic dimension $d$ of $\supp(\mu)$, which is much faster than the existing rate $\tilde{O}(n^{-2\beta / (2\beta + D)})$ \citep{schmidt2017nonparametric} without low dimensionality.
We note that the order of parameters $(W, L, B)$ is not affected by $D$; however, it does depend on $d$.
Moreover, using DNNs for estimation requires $\mO(1)$ layers to achieve the desired rate based on our approximation technique presented in Theorem \ref{thm:approx_mink}.

\vspace{\baselineskip}

\noindent \textbf{Proof Outline of Theorem \ref{thm:gen_mink}}:
First, we decompose the empirical loss into two terms, which are analogous to the bias and variance.
Following the definition of $\hat{f}$ as \eqref{def:fhat}, a simple calculation leads to
\begin{equation*}
  \|\hat f - f_0\|_n^2 \leq \underbrace{ \norm{f - f_0}_n^2}_{=:T_B} + \underbrace{\frac{2}{n}\sum_{i=1} \xi_i (\hat f(X_i) - f(X_i))}_{:= T_V},
\end{equation*}
for any $f \in \mF(W, L, B)$.
We define an empirical norm as $\norm{f}_n^2:=n^{-1}\sum_{i=1}^n f(X_i)^2$.

The first term $T_B$, which is analogous to an approximation bias, is evaluated using an approximation power of $\mF(W, L, B)$.
We apply Theorem \ref{thm:approx_mink} and bound the term.
The second term, $T_V$, which describes the variance of the estimator, is evaluated using the technique of the empirical process theory \citep{van1996weak}.
By using the notion of the local Rademacher complexity and concentration inequalities \citep{gine2016mathematical}, we bound $T_V$ by an integrated covering number of $\mF(W, L, B)$.
Further, we derive a bound for the covering number using the parameters $(W, L, B)$; thus, we can evaluate $T_V$ in terms of the parameters.
By combining the results for $T_B$ and $T_V$ and selecting proper values for $(W, L, B)$, we obtain the claimed result.

\subsection{Minimax Optimal Rate with Low Dimensionality}

We prove the optimality of the obtained rate in Theorem \ref{thm:gen_mink} by deriving the minimax error of the estimation problem.
To this end, we consider a probability measure $\mu$ with $\minkowski \supp(\mu) \leq d$.
Then, we obtain the following minimax lower bound.
\begin{theorem}[Minimax Rate with Low Dimensionality] \label{thm:minimax}
Let $\mP_d$ be a set of probability measures on $[0, 1]^D$ satisfying $\minkowski \supp(\mu) \leq d$.
Then, there exists a constant $C' > 0$ such that
\begin{align}
    \inf_{\check{f}} \sup_{(f_0, \mu) \in \mH(\beta, [0, 1]^D, M) \times \mP_d} \|\check{f} - f_0\|_{L^2(\mu)}^2 \geq C' n^{-2\beta / (2\beta + d)}, \label{ineq:minimax}
\end{align}
where $\check{f}$ is any arbitrary estimator for $f_0$.
\end{theorem}
That is, any estimator provides an error $\Omega (n^{-2\beta / (2\beta + d)})$ in a worst case scenario; therefore, it is regarded as a theoretical limit of efficiency.
Since the rate in Theorem \ref{thm:gen_mink} corresponds to the rate up to logarithmic factors, our rate almost achieves the minimax optimality.

\section{Comparison with Related Studies} \label{sec:compare}

\subsection{Nonparametric Analysis for DNNs}

\begin{table*}[htbp]
\vskip 0.15in
\begin{center}
\begin{small}
\begin{sc}
\begin{tabular}{cc|cc}
\toprule
 \multicolumn{2}{c|}{Setting} & \multicolumn{2}{c}{Error} \\
 $f_0$ & $\mu$ & Approximation & Estimation \\
\midrule
 H\"older / Sobolev &  & $\tilde{O}(W^{-\beta/D})$ & $\tilde{O}(n^{-2\beta / (2\beta + D)})$ \\
 Barron &  & $\tilde{O}(W^{-1/2})$ & $\tilde{O}(n^{-1})$ \\
 Mixed Smooth &  & $\tilde{O}(W^{-\gamma})$ & $\tilde{O}(n^{-2\gamma / (2\gamma + 1)})$ \\
 \textbf{H\"older} & \textbf{$d$-dimensional}   & $O(W^{-\beta/d})$ & $\tilde{O}(n^{-2\beta / (2\beta + d)})$ \\
\bottomrule
\end{tabular}
\end{sc}
\end{small}
\end{center}
\vskip -0.1in
\caption{Comparison of the derived rates of approximation and generalization errors with a non-parametric class of target functions. $W$ denotes a number of parameters in DNNs, and $n$ is the number of observations. $D$ is the dimension of $X$, $\beta$ is the smoothness of $f_0$, $\gamma$ is the index of mixed smoothness, and $d$ is an intrinsic dimension of $X$.
\label{tab:rates_compare}}
\end{table*}

Many studies have investigated the approximation and estimation performance of DNNs and some \citep{yarotsky2017error,schmidt2017nonparametric} clarify the performance of DNNs with ReLU activations when $f_0$ is in the H\"older space.
They show that this performance is $\tilde{O}(W^{-\beta/D})$ for approximation and $\tilde{O}(n^{-2\beta/(2\beta + D)})$ for estimation.
Other studies \citep{montanelli2017deep,suzuki2018adaptivity} consider a different functional class with \textit{mixed smoothness} for $f_0$ and then obtain a novel convergence rate that depends on its particular smoothness index $\gamma$.
Additionally, other studies have investigated more specific structures of $f_0$.
If we can decompose $f_0$ into a composition of feature maps, the convergence rate depends on the dimensionality of the feature space \citep{petersen2018optimal}.
As a close analog to this study, the DNNs’ error converges faster when $f_0$ is associated with a manifold structure \citep{schmidt2019deep}.
{\bc
\citet{bach2017breaking} derives an error for a class of Lipschitz-continuous functions, which is considered as a special case of the H\"older space with $\beta = 1$.

A large number of research have been conducted to achieve faster convergence rates by introducing specific structures.
\citet{bach2017breaking} showed that $D$-independent convergence rates can be achieved when $f_0$ has a parametric structure such as a general additive model or a single index model.
Similarly, when $f_0$ has the form of a generalized hierarchical version of a single index model \citep{bauer2019deep} or a form with multivariate adaptive regression splines (MARS) \citep{kohler2019deep}, we can obtain a faster convergence rate with DNNs.
Although these rates are fast, it is unlikely that $f_0$ has such specific parametric model structures in practice.
As a non-parametric attempt to obtain a faster convergence rate, a classical approach \citep{barron1993universal,barron1994approximation} considers a restricted functional class (referred to as the \textit{Barron class} in this study) for $f_0$, and achieving a very fast rate: $\tilde{O}(W^{-1/2})$ for approximation, and $\tilde{O}(n^{-1})$ for estimation.
The Barron class is non-parametric and has no model constraints, but it has constraints on differentiability through spectral conditions.
Given that the class requires higher differentiability when $D$ is large, a more flexible functional class is required.
}

\begin{figure}
  \begin{center}
    \includegraphics[width=0.9\textwidth]{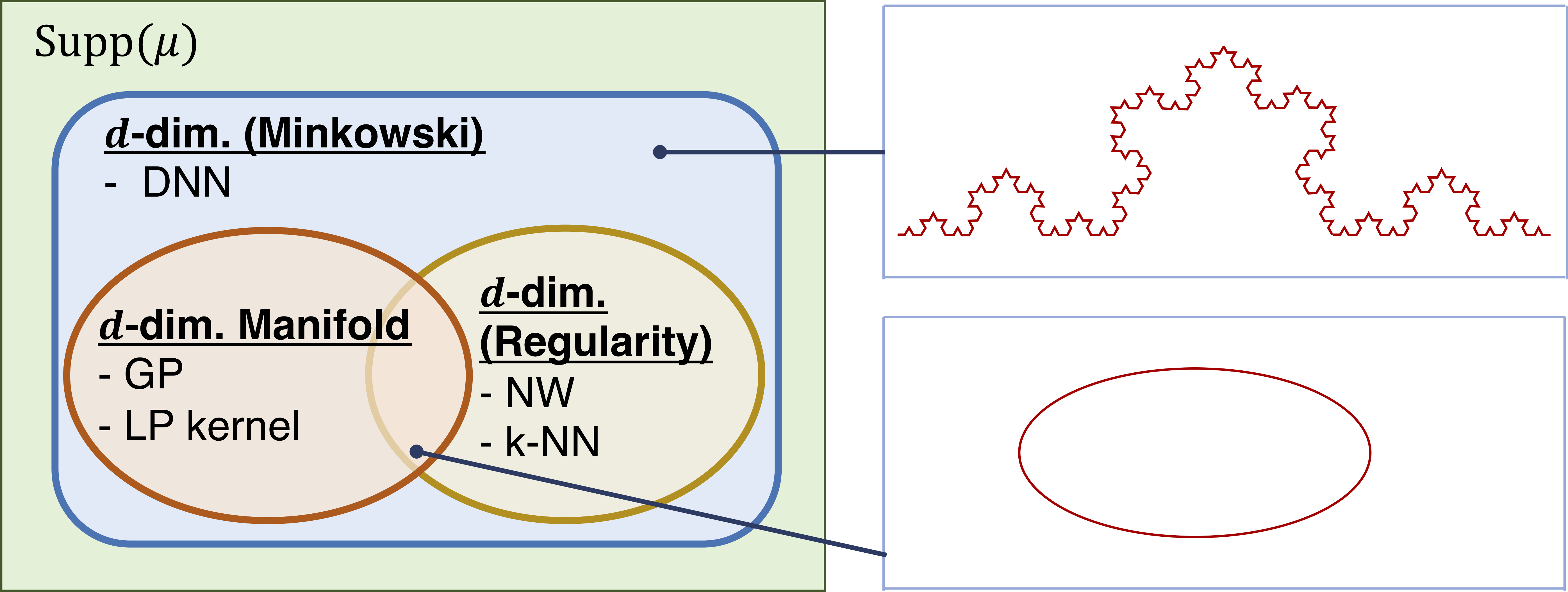}
  \end{center}
  \caption{[Left] Various low dimensional sets as $\supp(\mu)$, and corresponding regression methods that can obtain the optimal rate on each of the sets. The Minkowski dimension can describe a wider class of low dimensional sets. [Right] The top right is the Koch curve, which is a low dimensional set in terms of the Minkowski dimension. The bottom right is an ellipse representing a smooth manifold. It is also a low dimensional set in terms of the Minkowski and the regularity dimension. \label{fig:compare}}
\end{figure}

Our study assumes that $f_0$ is an element of the H\"older space, and that $\mu$ has an intrinsic low dimensional structure; that is, $\supp(\mu)$ is $d$-dimensional in the Minkowski or the manifold sense.
{\bc
The H\"older class allows the smoothness of the function to be determined separately of $D$.
Thus, we can investigate the effects of $d$ and $D$ independently.
}
We obtain the approximation rate $O(W^{-\beta / d})$ and the estimation rate $\tilde{O}(n^{-2\beta / (2\beta + d)})$.
Because $d$ is much less than $D$ empirically (as mentioned in Section \ref{section:empirical}), the rate can alleviate the curse of dimensionality caused by a large $D$.
To the best of our knowledge, this is the first study proving that errors in DNNs converge faster with data having general intrinsic low dimensionality.

\subsection{Other Adaptive Methods with Intrinsic Low Dimensionality}

Except for DNNs, several nonparametric estimators can obtain a convergence rate that is adaptive to the intrinsic dimension of a distribution $d$.
The local polynomial kernel (LP kernel) regression \citep{bickel2007local} and the Gaussian process (GP) regression \citep{yang2016bayesian} can achieve the rate $\mO(n^{-2\beta/(2\beta + d')})$, where $\supp(\mu)$ is a $d'$-dimensional manifold with $\beta = 2$.
Similarly, the $k$-nearest neighbor (k-NN) regression \citep{kpotufe2011k} and the Nadaraya-Watson (NW) kernel regression \citep{kpotufe2013adaptivity} can achieve the rate with $d'$, when $\mu$ has a \textit{regularity dimension} $d'$, that is less general than the Minkowski dimension (Lemma \ref{lem:mink_regular} in the supplementary material).

We show that DNNs with finite layers can obtain the fast convergence rate of $\mO(n^{-2\beta/(2\beta + d)})$ over a broader scope of cases compared to the existing adaptive methods.
Theorem \ref{thm:gen_mink} indicates that DNNs with $L=\mO(1)$ can obtain the rate when $\dim_M\supp(\mu) < d$, which is less restrictive than the settings with manifolds and regularity dimensions.
Intuitively, DNNs can obtain a fast adaptive rate even when $\supp(\mu)$ does not have a smooth structure such as that of manifolds.
Figure \ref{fig:compare} presents an overview of the results.

\section{Simulation} \label{sec:numerical}

\subsection{Estimation by DNNs with Different $d$} \label{sec:exp_dnn}

\begin{wrapfigure}[20]{r}[5mm]{0.44\hsize}
\centering
    \includegraphics[clip,width=0.95\hsize]{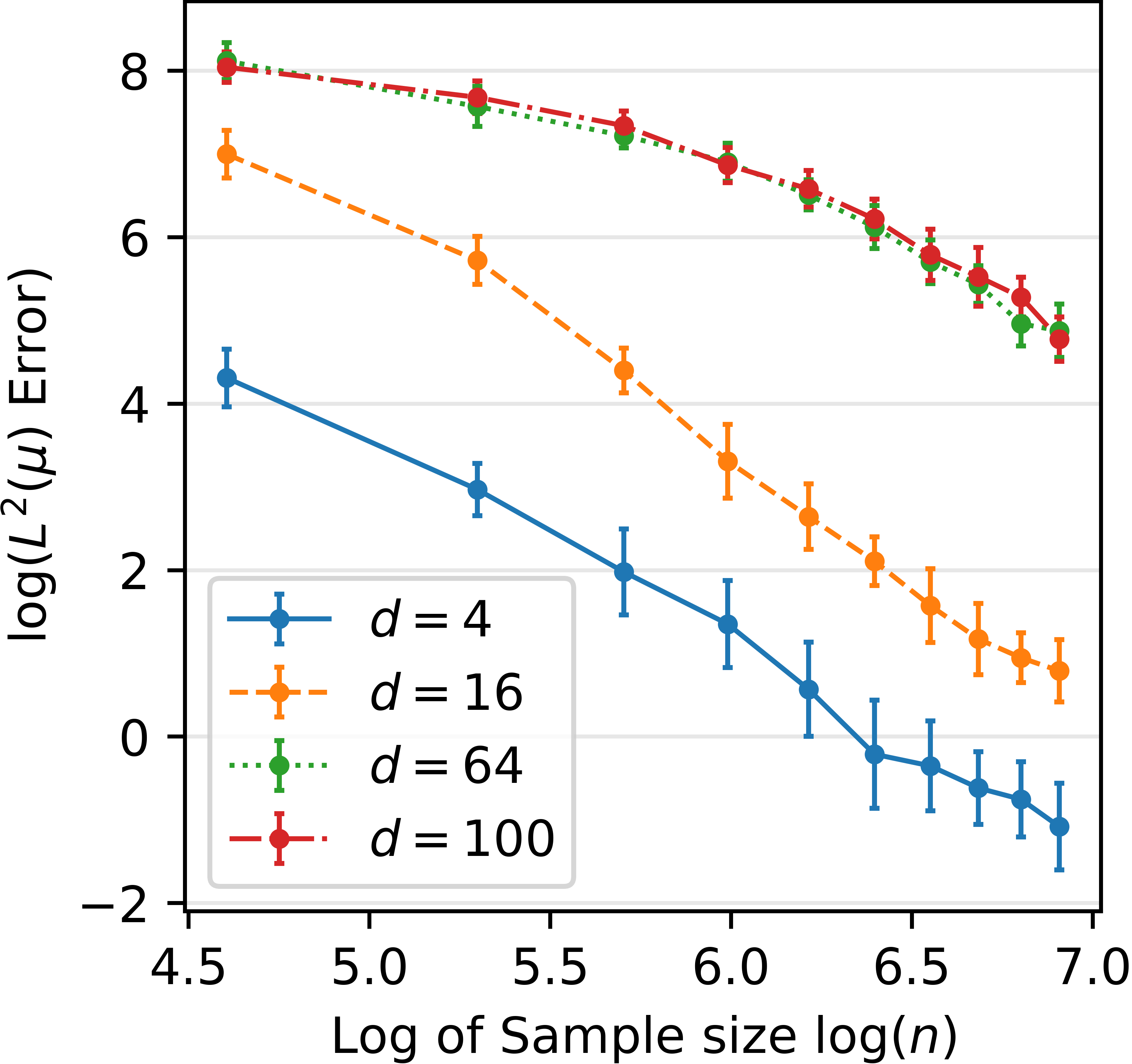}
    \caption{Simulated generalization errors of DNNs with $D=128$ and $d \in \{4,16,64,100\}$.
    The error bars show the standard deviation from the replication.
    \label{fig:risk_dnn}}
\end{wrapfigure}

We calculate the generalization errors of DNNs using synthetic data.
We set the true function as
$f_0(x) := (D-1)^{-1}\sum_{i=1}^{D-1} x_i x_{i+1} + D^{-1} \sum_{i=1}^D 2\sin(2 \pi x_i) \1_{\{x_i \leq 0.5\}}+D^{-1} \sum_{i=1}^D (4\pi(\sqrt{2} - 1)^{-1} (x_i - 2^{-1/2})^2 - \pi(\sqrt{2} - 1))\1_{\{x_i > 0.5\}}$, which belongs to $\mH(\beta, [0, 1]^D)$ with $\beta = 2$.
We set $\mu$ as a uniform measure on a $d$-dimensional sphere embedded in $[0,1]^D$, and also set a noise $\xi_i$ as a Gaussian variable with zero mean and variance $\sigma^2 = 0.1$.
We generate $n$ pairs of $(X_i, Y_i)$ from the regression model \eqref{eq:model} and learn the estimator \eqref{def:fhat}.
For the learning process, a DNN architecture with four layers and the ReLU activation function are employed, and each layer has $D$ units except the output layer.
For optimization, we employ Adam \citep{kingma2015adam} with the following hyper-parameters; $0.001$ learning rate and $(\beta_1,\beta_2) = (0.9,0.999)$.

We set the nominal dimension as $D = 128$ and consider different numbers of samples $n \in \qty{100,200,..., 1000}$ and intrinsic dimensions $d \in \{ 4, 16, 64, 100\}$.
We measure the generalization errors using validation data in terms of the $L^2(\mu)$-norm.
We replicate the learning procedure $100$ times with different initial weights for the parameters of neural networks from a standard normal distribution.

We plot the generalization errors in log against $\log n$ in Figure \ref{fig:risk_dnn}.
The slope of the curve corresponds to the convergence rate of the errors as the figure is double logarithmic.
From the results, we observe the following two findings:
(i) The error is lower with a fewer $d$, and
(ii) The convergence rates with $d \in \{4,16\}$ are faster than those with $d \in \{ 64,100\}$.

\begin{figure}
\centering
\begin{minipage}{0.49\hsize}
    \centering
    \includegraphics[clip,width=0.95\hsize]{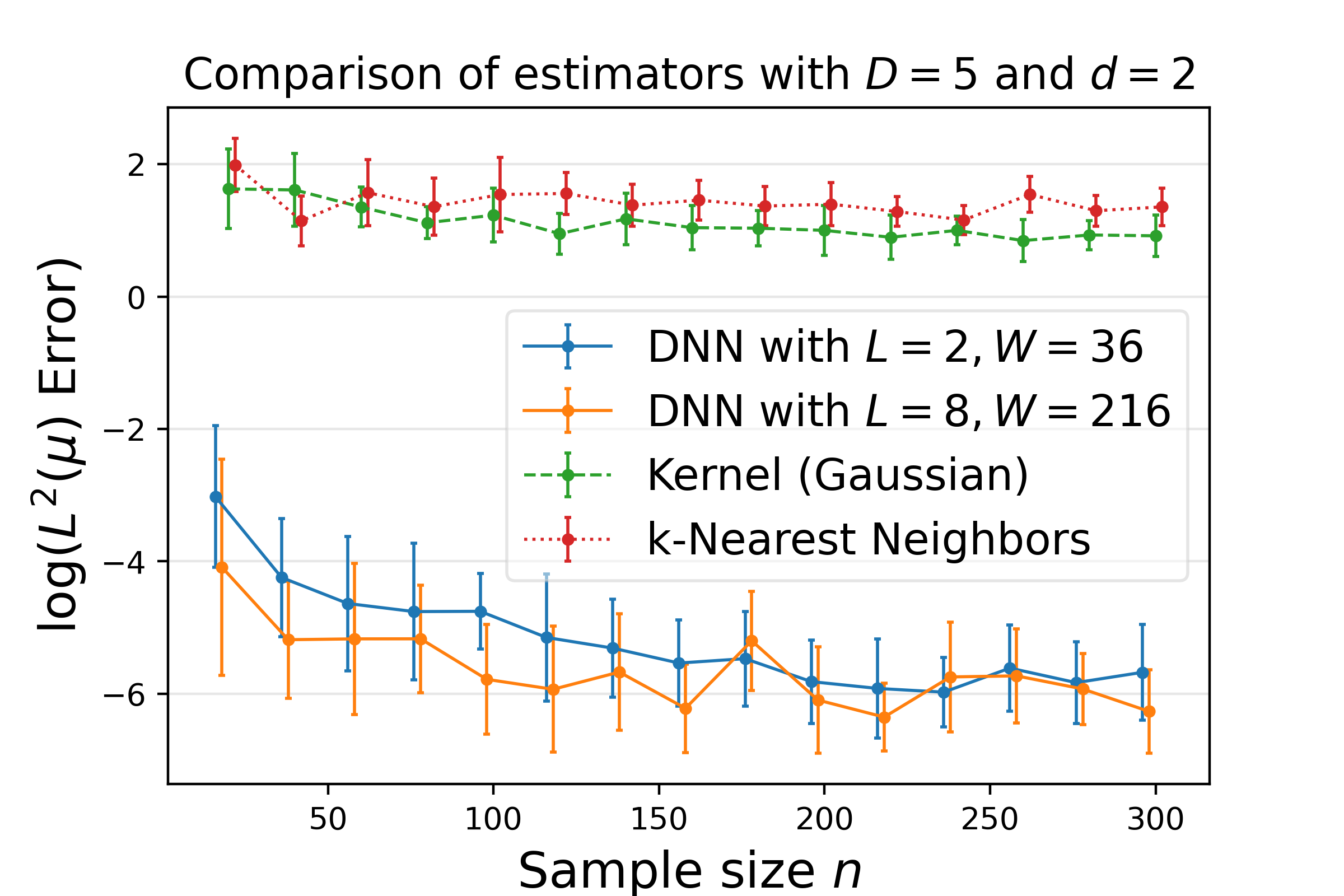}
\end{minipage}
\begin{minipage}{0.49\hsize}
    \centering
  \includegraphics[clip,width=0.95\hsize]{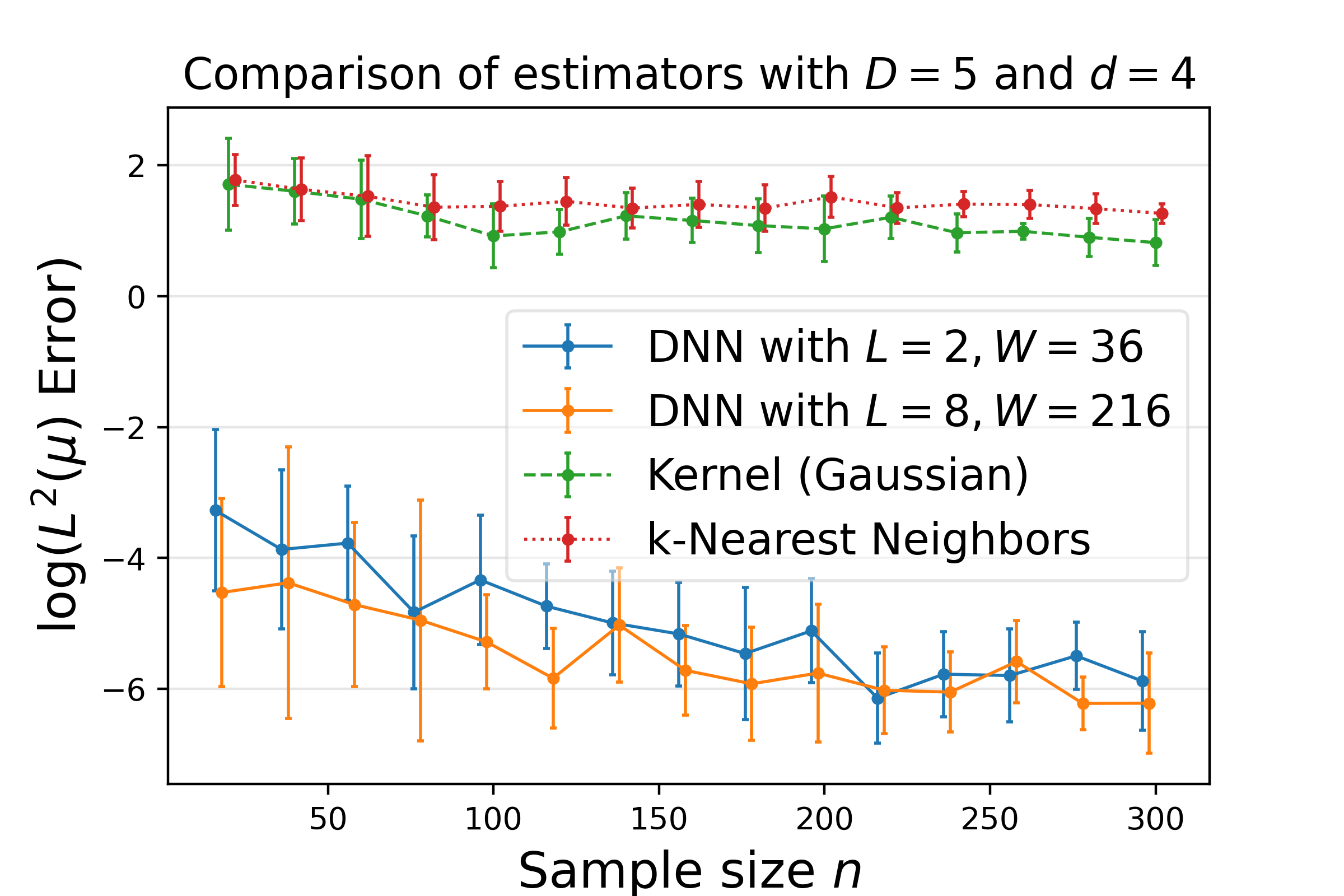}
\end{minipage}
    \caption{Simulated generalization errors for NW with Gaussian kernel, k-NN regression, and DNNs ($2$ layers and $8$ layers).
    The error bars shows the standard deviation of the $10$ replication.
    The left panel shows the case with $D=5$ and $d=2$, and the right panel is for $D=5$ and $d=4$.
    \label{fig:comparison_gen}}
\end{figure}

\subsection{Comparison with the Other Estimators}

We compare the performances of DNNs with existing methods, such as $k$-NN method and the NW kernel method.
We note that they can achieve a rate whose exponent depends on $d$.

{\bc
We set the true function as $f_0(x) := (1/D) \sum_{i=1}^{D} x_i^2\1_\qty{x_i \leq 0.5} + (-x_i + 3/4) \1_\qty{x_i > 0.5}$, which belongs to $\mH(\beta, [0, 1]^D)$ with $\beta = 1$.
Let $\mu$ be the uniform measure on 
a union of $d$-dimensional $\ell^{1/2}$ ball and $\ell^2$ ball embedded in $[0,1]^D$.
We note that the support of $\mu$ is not a smooth manifold.
We set $n \in \{20,40,...,300\}$, and also consider two configurations $(D,d) = (5,2)$ and $(D,d) = (5,4)$.
For each sample size, we replicate the estimation $10$ times with different initial weights from a standard normal distribution.
The learning procedure of DNNs is the same as that described in Section \ref{sec:exp_dnn}.
For $k$-NN, its hyper-parameter $k$ is selected from $[50]$.
Fir the NW kernel method, we employ a Gaussian kernel whose bandwidth is selected from $\qty{0.10, 0.11, \dots, 1.00}$.
We select all the hyper-parameters based on cross-validation.

We plot the simulated generalization error with the validation data and the other methods based on $n$ in Figure \ref{fig:comparison_gen}.
The results indicate that DNNs outperform the other estimators.
Since $\supp(\mu)$ is not a smooth manifold, this is likely to affect the dominance observed for the DNNs.
}

\subsection{Real Data Analysis}

We compare the performance of DNNs using the modified National Institute of Standards and Technology (MNIST) dataset.
The dataset contains 784-dimensional grayscale images of handwritten digits.

For the experiment, we contaminate the data with $d$-dimensional Gaussian noise using $d \in \qty{0, 25, 50, 75, 100}$ with variance $0.01$.
We set $n \in \qty{100, 200, 300, 400, 500}$.
Considering that the task is classification, we employ the soft-max activation function in the last layer and measure the error based on the $\|\cdot \|_{L^2}$-norm.
We replicate the setting $10$ times, and discard two replications with the first and second-largest test errors to eliminate the effect of the difficulty involved with non-convex optimization using DNNs.
All the other settings inherit those of Section \ref{sec:exp_dnn}.

\begin{table*}[htbp]
\vskip 0.15in
\begin{center}
\begin{small}
\begin{sc}
\begin{tabular}{c|ccccc}
\toprule
$d$ & 0 & 25 & 50 & 75 & 100 \\
\midrule
Convergence Rate & -0.20 & -0.19 & -0.17 & -0.18 & -0.16\\
\bottomrule
\end{tabular}
\end{sc}
\end{small}
\end{center}
\vskip -0.1in
\caption{Estimated convergence rate with the MNIST data.
$d$ denotes an intrinsic dimension of the contaminated Gaussian noise.
The convergence rate is estimated by regressing the logarithm of test error on $\log n$.
\label{tab:mnist_dnn}}
\end{table*}

Table \ref{tab:mnist_dnn} shows the convergence rates of the generalization error for each value of $d$.
The rate is estimated by using the least square method for the logarithm of the test error versus the logarithm of the sample size.
The result indicates that the convergence gets slower as $d$ increases.

\section{Conclusion} \label{sec:conclusion}

We theoretically elucidate that the intrinsic low dimensionality of data mainly determines the performance of deep learning.
To show the result, we introduce the notion of the Minkowski dimension and then derive the rates of approximation and generalization errors, which only depend on the intrinsic dimension and are independent of the large nominal dimension.
Additionally, we find that DNNs can achieve the convergence rate over a broader class of data than several conventional methods.
Our results provided evidence of the inherent convergence advantage of deep learning over other models with respect to data with low intrinsic dimension.

\acks{We would like to thank the editor for valuable comments and suggestions. We also acknowledge support for this project from JSPS KAKENHI (18K18114, 18K19793 and 20J21717) and JST Presto.}


\appendix
\section{Supportive Discussion}

\subsection{Additional Notation and Definition}
A closed ball in $\R^D$ with its center $x$ and radius $r$ with norm $\norm{\cdot}$ is described as $\overline{B}^D(x,r) := \{x' \in \R^D \mid \|x-x'\| \leq r\}$.
An open ball is similarly defined as $B^D(x, r)$.
A closed ball in $\R^D$ with its center $x$ and radius $r$ with the $\ell^p$-norm is described as $\overline{B}^D_p(x,r)$.
An open version of the ball is $B^D_p$.
For a set $\Omega$, $\1_\Omega(\cdot)$ is an indicator function such that $\1_\Omega(x)=1$ if $x \in \Omega$, and $\1_\Omega(x)=0$ otherwise.
We write the set of non-negative real numbers as  $\R_\geq$.
For sequences $\{a_n\}_n$ and $\{b_n\}_n$, $a_n \lesssim b_n$ denotes $a_n \leq C b_n$ with a finite constant $C>0$ for all $n$.
With $\varepsilon > 0$, $\mN_2(\Omega, \varepsilon)$ is a smallest number of $\varepsilon$-balls, which cover $\Omega$ in terms of $\|\cdot\|_{2}$.
Rigorously, a support of a probability measure $\nu$ on a set $\mX$ is defined as $\supp(\nu) \defeq \qty{x \in \mX \mid V \in \mN_x \Rightarrow \nu(V) > 0}$, where $\mN_x$ is a set of open neighborhoods with its center $x \in \mX$.
We define a standard distance $d(A, B)$ of two sets $A$ and $B$ by
\begin{equation*}
    d(A, B) := \inf\qty{\norm{x - y} \mid x \in A, y \in B}.
\end{equation*}

\subsection{Other Notions for Dimensionality} \label{sec:dimensions}

We show a relation between the Minkowski dimension and the dimension of manifolds.
The notion of \textit{manifolds} is common for analyzing low dimensionality of data \citep{belkin2003laplacian,niyogi2008finding,genovese2012minimax}.
The Minkowski dimension can describe the dimensionality of manifolds.


\begin{lemma} \label{lem:manifolds}
    Let $\mM$ be a compact $d$-dimensional manifold in $[0,1]^D$.
  Assume $\mM = \bigcup_{k=1}^K \mM_k \subset [0, 1]^D$ for $K \in \N$. Also assume that for any $1 \leq k \leq K$, there exists an onto and continuously differentiable map $\psi_k : [0, 1]^{d_k} \to \mM_k$ each of which has the input dimension $d_k \in \N$.
  Then, $\minkowski \mM \leq \max_{1\leq k \leq K} d_k$.
\end{lemma}

\noindent \textbf{Proof of Lemma \ref{lem:manifolds}}
We first assume that the statement holds with $K=1$.
Then, we investigate the case with general $K$.

Suppose the lemma is correct for $K = 1$.
Take $d^*_k > d_k$. Since $\minkowski \mM_k < d^*_k$, there exists a constant $C > 0$ such that for any $\varepsilon > 0$, existing a finite set $F^k_\varepsilon \subset [0, 1]^D$ satisfies the followings:
  \begin{enumerate}
    \item $\mM_k \subset \bigcup_{x \in F^k_\varepsilon} B^D_2(x, \varepsilon)$,
    \item $\card(F_\varepsilon^k) \leq C_k \varepsilon^{-d^*_k}$.
  \end{enumerate}
  Let $F_\varepsilon \defeq \bigcup_{k=1}^K F_\varepsilon^k$, then we have $\mM \subset \bigcup_{x \in F_\varepsilon} B^D_2(x, \varepsilon)$ and
  \begin{align*}
      \mN_2(\mM, \varepsilon) \leq \card(F_\varepsilon) \leq \qty(\sum_{k=1}^K C_k) \varepsilon^{-\max_{k} d^*_k}.
  \end{align*}
  For $d^* > d \defeq \max_k d_k$, we choose $d^*_k > d_k$ such that $\max_k d^*_k < d^*$ holds.
  Then, it yields $\limsup_{\varepsilon \downarrow 0} \mN_2(\mM, \varepsilon) \varepsilon^{d^*} = 0$.
  So the proof is reduced to the case of $K = 1$.

  We investigate the case $K=1$.
  For brevity, we omit the subscript $k$ and write $\psi=(\psi_1,...,\psi_D)$.
  Recall that $\psi_i$ is continuously differentiable. We also define
  \begin{align*}
      L_i \defeq \max_{x \in [0,1]^D} \sqrt{\sum_{j=1}^d \abs{\partial \psi_i'(x)/ \partial x_j}^2}.
  \end{align*}
  Applying the mean-value theorem to $\psi_i$ yields $\abs{\psi_i(z) - \psi_i(w)} \leq L_i\norm{z - w}_2$. By the Lipschitz continuity of $\psi = (\psi_1, \dots, \psi_D)$, for any $z, w$, $\norm{\psi(z) - \psi(w)}_2 \leq \sqrt{D} L \norm{z - w}_2$ where $L \defeq \max_i L_i$.
  Fix any $\varepsilon > 0$ and $d \in \N$. Recall that there exists a constant $C > 0$ so that for any $\delta > 0$, an existing finite set $F_{\delta} \subset [0, 1]^d$ (see Example 27.1 in \citet{shalev2014understanding}) satisfies

  \begin{enumerate}
    \item $\card(F_{\delta}) \leq C \delta^{-d}$,
    \item $[0, 1]^d \subset \bigcup_{y \in F_{\delta}} B^d_2(y, \delta)$.
  \end{enumerate}
  Choosing $\delta = \varepsilon/(\sqrt{D}L)$ yields
  \begin{align*}
    \mM &\subset \psi\qty(\bigcup_{y \in F_{\delta}} B^d_2(y, \delta) \cap [0, 1]^d)
    \subset \bigcup_{y \in F_{\delta}} \psi\qty(B^d_2(y, \delta) \cap [0, 1]^d)
    \subset \bigcup_{y \in F_{\delta}} B^D_2(\psi(y), \varepsilon)
  \end{align*}
  where the last inclusion follows from the Lipschitz continuity of $\psi$.
  Since we have
  \begin{align*}
      \card(F_{\delta}) \leq C(\sqrt{D}L)^d \varepsilon^{-d},
  \end{align*}
  we obtain the conclusion.
\qed

Additionally, we explain the notion of a \textit{doubling measure}, which is an alternative way to describe an intrinsic dimension.
It is employed in several studies \citep{kpotufe2011k,kpotufe2013adaptivity}.
\begin{definition}[Doubling Measure]
  A probability measure $\nu$ on $\mX$ is called a doubling measure, if there exists a constant $C > 0$ such as
  \begin{align*}
    \nu(\overline{B}^D(x,2r) \cap \mX) \leq C \nu(\overline{B}^D(x, r) \cap \mX),
  \end{align*}
  for all $x \in \supp(\nu)$ and $r > 0$.
\end{definition}
Then, we can define a dimensionality by the regularity property of doubling measures.
\begin{definition}[Regularity Dimension]
  For a doubling measure $\nu$, the (upper) regularity dimension $\regularity \nu$ is defined by the infimum of $d^* > 0$ such that there exists a constant $C_\nu > 0$ satisfying
  \begin{align*}
    \frac{\nu(\overline{B}^D(x,r) \cap \mX)}{\nu(\overline{B}^D(x,\varepsilon r) \cap \mX)} \leq C_\nu \varepsilon^{-d^*},
  \end{align*}
  for all $x \in \supp(\nu)$, $\varepsilon \in (0, 1)$ and $r > 0$.
\end{definition}

There also exists a relation between the \textit{Minkowski dimension} and the \textit{regularity dimension}.
Intuitively, when $\mX \subset \R^D$, a measure $\nu$ with $\regularity \nu = d$ behaves as if a domain of $\nu$ is $\R^d$, as shown in Figure \ref{fig:dim}.
The regularity dimension of $\nu$ can evaluate the Minkowski dimension of $\supp(\nu)$.
\begin{lemma}[Lemma 3.4 in \citet{kaenmaki2013dimensions}] \label{lem:mink_regular}
  Let $(\mX, \mB, \nu)$ be a probability space with $\mX \subset \R^D$ is bounded.
  Suppose $\nu$ is a doubling measure. Then,
  \begin{equation*}
    \minkowski \supp(\nu) \leq \regularity \nu.
  \end{equation*}
\end{lemma}





\section{Proof of Main Results}

\subsection{About Theorem \ref{thm:approx_mink}}

Our proof strategy is an extended version of those of the previous studies \citep{yarotsky2017error,petersen2018optimal}.
The previous studies employ a simultaneous approximation of Taylor polynomials multiplied by approximated indicator functions for each disjoint hypercube.
However, their approach fails with the concentrated measure of covariates.
To avoid the problem, we develop a novel way to unite the approximators by the disjoint subsets and the max operation as described in the proof outlined in Section \ref{sec:approx}. 

\subsubsection{Sub-Neural Networks as Preparation}

Before a central part of the proof, for convenience, we define several sub-neural networks: (i) concatenation of neural networks, (ii) parallelization of neural networks, (iii) parallelization of neural networks with different inputs, (iv) an approximated identity function, and (v) a max function.

\textbf{(i) Concatenation of Neural Networks}:
Given two neural networks $\Phi^1$ and $\Phi^2$, we aim to construct a network $\Phi$ such that $R(\Phi) = R(\Phi^2) \circ R(\Phi^1)$, which is possible because ReLU activation function has the property $\rho(x) - \rho(-x) = x$.
Write $\Phi^1 = ((A^1_{L_1}, b^1_{L_1}), \dots, (A^1_{1}, b^1_{1}))$ and $\Phi^2 = ((A^2_{L_1}, b^2_{L_1}), \dots, (A^2_{1}, b^2_{1}))$.
We define parameter matrices and vectors as
\begin{equation*}
  \tilde A^2_1 \defeq \qty(\begin{array}{c}
    A^2_1\\
    -A^2_1
  \end{array}),\,
  \tilde b^2_1 \defeq \qty(\begin{array}{c}
    b^2_1\\
    -b^2_1
  \end{array}),\,
  \tilde A^1_{L_1} \defeq \qty(A^1_{L_1} \, -A^1_{L_1}).
\end{equation*}
Then, the concatenation of neural networks $\Phi^1$ and $\Phi^2$ is defined as
\begin{align*}
  \Phi^2 \odot \Phi^1 &\defeq \bigl(
    (A^2_{L_2}, b^2_{L_2}), \dots, (A^2_{2}, b^2_{2}), (\tilde A^2_{1}, \tilde b^2_{1}), (\tilde A^1_{L_1}, b^1_{L_1}), (A^1_{L_1-1}, b^1_{L_1-1}), \dots, (A^1_{1}, b^1_{1}).
  \bigr)
\end{align*}
It is easy to show the following relations:
\begin{enumerate}
  \item $W(\Phi^2 \odot \Phi^1) \leq 2W(\Phi^2) + 2W(\Phi^1)$,
  \item $L(\Phi^2 \odot \Phi^1) = L(\Phi^2) + L(\Phi^1)$,
  \item $B(\Phi^2 \odot \Phi^1) = \max\qty{B(\Phi^2), B(\Phi^1)}$.
\end{enumerate}

By repeating the discussion, we obtain the concatenation of $k$ neural networks.
Some properties of the concatenation are summarized in the following remark.
\begin{remark}[Concatenation]\label{rem:concatenation}
  For any neural networks $\Phi^1, ..., \Phi^k$, we have the following inequalities:
  \begin{align*}
    W(\Phi^k \odot \cdots \odot \Phi^1) &\leq 2 \sum_{i=1}^k W(\Phi^i),\\
    L(\Phi^k \odot \cdots \odot \Phi^1) &= \sum_{i=1}^k L(\Phi^i),\\
    B(\Phi^k \odot \cdots \odot \Phi^1) &= \max_{1 \leq i \leq K} B(\Phi^i).
  \end{align*}
\end{remark}

\textbf{(ii) Parallelization of Neural Networks}:
We define the parallelization of multiple neural networks.
Let $\Phi^i = ((A^i_{L_i}, b^i_{L_i}), \dots, (A^i_1, b^i_1)),$ be neural networks with a $d_i$-dimensional input and an $m_i$-dimensional output.
When the input dimension $d_i = d$ for all $i$, we can define the parallelization of neural networks.

Suppose the number of layers of the network is the same for all $\Phi^i$,
Write ${\Phi^i}' = ((A^i_{L}, b^i_{L}), \dots, (A^i_1, b^i_1))$ and define parameters as
\begin{align*}
  A_{1} \defeq \begin{pmatrix}
    {A^1_{1}}^\top&
    {A^2_{1}}^\top&
    \cdots&
    {A^K_{1}}^\top
  \end{pmatrix}^\top, \mbox{~and~}
  b_{1} \defeq \begin{pmatrix}
    {b^1_{1}}^\top&
    \cdots&
    {b^K_{1}}^\top
  \end{pmatrix}^\top.
\end{align*}
Also, for $\ell \geq 2$, we define parameters as
\begin{align*}
  A_{\ell} \defeq \left(\begin{array}{cccc}
    A^1_{\ell} & O &\ldots & O\\
    O & A^2_{\ell} & \ldots & O\\
    \vdots & \vdots & \ddots & \vdots\\
    O & O & \ldots & A^K_{\ell}
  \end{array}\right), \mbox{~and~}
  b_{\ell} \defeq \left(\begin{array}{c}
    b^1_{\ell}\\
    \vdots\\
    b^K_{\ell}
  \end{array}\right).
\end{align*}
Then, we define the parallelization of networks $(\Phi^i)_{i=1}^K$ with $d$-dimensional input and $\sum_{i=1}^K m_i \eqdef m$-dimensional output as
\begin{equation*}
  [\Phi^1, \Phi^2, \dots, \Phi^K] \defeq ((A_{L}, b_{L}), \dots, (A_1, b_1)).
\end{equation*}
Combined with the previous results, we can show the following result.
\begin{remark}[Parallelization]\label{rem:parallelization}
Let $\Phi^i = ((A^i_{L_i}, b^i_{L_i}), \dots, (A^i_1, b^i_1))$ be the neural network with $d$-dimensional input, $m_i$-dimensional output and the same number of layers $L$ for $i \in \qty{1, \dots, K}$.
Then, we obtain the followings:
\begin{align*}
  W([\Phi^1, \Phi^2, \dots, \Phi^K]) &= \sum_{i=1}^K W(\Phi^i),\\
  L([\Phi^1, \Phi^2, \dots, \Phi^K]) &= L,\\
  B([\Phi^1, \Phi^2, \dots, \Phi^K]) &= \max_{1\leq i \leq K} B(\Phi^i).
\end{align*}
\end{remark}

\textbf{(iii) Parallelization of Neural Networks with Different Input}:
We define the parallelization of multiple neural networks without sharing the input.
Under this case, the networks share the number of layers only.
Let $\Phi^i = ((A^i_{L}, b^i_{L}), \dots, (A^i_1, b^i_1))$ be a neural network with a $d_i$-dimensional input and an $m_i$-dimensional output.
Define $m \defeq \sum_{i=1}^K m_i$ and $d \defeq \sum_{i=1}^K d_i$. We construct a neural network version of a function $\R^d \to \R^m$
\begin{equation*}
   \qty((x_{1, 1}, \dots x_{1, d_1}), \dots, (x_{K, 1}, \dots, x_{K, d_K}))^\top \mapsto \begin{pmatrix} R(\Phi_1)(x_{1, 1}, \dots, x_{1, d_1})\\
    \vdots\\
    R(\Phi_K)(x_{K, 1}, \dots, x_{K, d_K}) \end{pmatrix}.
\end{equation*}
Write $\Phi^i = ((A^i_{L}, b^i_{L}), \dots, (A^i_1, b^i_1))$ and we define parameters as
\begin{align*}
  A_{\ell} \defeq \left(\begin{array}{cccc}
    A^1_\ell & O &\ldots & O\\
    O & A^2_\ell & \ldots & O\\
    \vdots & \vdots & \ddots & \vdots\\
    O & O & \ldots & A^K_\ell
  \end{array}\right),\mbox{~and~}
  b_{\ell} \defeq \left(\begin{array}{c}
    b^1_\ell\\
    \vdots\\
    b^K_\ell
  \end{array}\right),
\end{align*}
We define the parallelization of networks $(\Phi^i)_{i=1}^K$ with $d$-dimensional input and $m$-dimensional output as
\begin{equation*}
  \langle\Phi^1, \Phi^2, \dots, \Phi^K\rangle \defeq \qty((A_{L}, b_{L}), \dots, (A_1, b_1)).
\end{equation*}
We obtain the following remark:
\begin{remark}[Parallelization with Different Input]\label{rem:parallelization_for_different_input}
Let $\Phi^i = ((A^i_{L_i}, b^i_{L_i}), \dots, (A^i_1, b^i_1))$, $i \in [K]$ be neural networks with $d_i$-dimensional input and $m_i$-dimensional output with the same number of layers $L$.
Then, we obtain the following relations:
\begin{align*}
  W\qty(\langle\Phi^1, \Phi^2, \dots, \Phi^K\rangle) &= \sum_{i=1}^K W(\Phi_i),\\
  L\qty(\langle\Phi^1, \Phi^2, \dots, \Phi^K\rangle) &= L,\\
  B\qty(\langle\Phi^1, \Phi^2, \dots, \Phi^K\rangle) &= \max_{1\leq i \leq K} B(\Phi^i).
\end{align*}
\end{remark}

\textbf{(iv) Identity Function}:
We define a neural network that approximates an identity function.
For $L \geq 2$, let $\Phi^{Id}_{D, L} : \R^D \to \R^D$ be a neural network of identity function
\begin{equation*}
  \Phi^{Id}_{D, L} \defeq \qty( \qty( \qty(I_D \, -I_D), 0 ), \underbrace{ (I_{2D}, 0) , \dots, (I_{2D}, 0)}_{L-2 \text{ times}}, \qty( \qty(\begin{array}{c}I_D\\-I_D\end{array}), 0 )),
\end{equation*}
where $I_p \in \R^{p \times p}$ is the identity matrix.
For $L = 1$, let $\Phi^{Id}_{D, L} = ((I_D, 0))$.
Then its realization is the identity function $\R^p \ni x \mapsto x$.
We can see $W(\Phi^{Id}_{D, L}) = 2DL$, $L(\Phi^{Id}_{D, L}) = L$ and $B(\Phi^{Id}_{D, L}) = 1$ for all $L \geq 2$.

\textbf{(v) Max function}:
We define a neural network that works as a max function.
We implement $\max: \R_{\geq}^s \to \R$ by a neural network.
Let $t = \ceil{\log_2 s}$. Note that $\log_2 s \leq t < \log_2 s + 1$ and $s \leq 2^t < 2s$.
With the ReLU activation function, the max function $(x_1, x_2) \mapsto \max(x_1, x_2)$ for non-negative inputs can be easily implemented by a ReLU neural network
\begin{equation*}
    \Phi^{\max, 2} \defeq \qty(((1, 1), 0), \qty(\begin{pmatrix} 1 & 0\\ -1 & 1 \end{pmatrix}, \begin{pmatrix} 0\\ 0 \end{pmatrix})),
\end{equation*}
due to the identity $\max(x_1, x_2) = \rho(x_2 - x_1) + \rho(x_1)$ for $x_1, x_2 \geq 0$.

Next, we extend $\Phi^{\max, 2}$ to take multiple inputs.
Since $2^t = s$ does not always hold, we need a following dummy network to make $\Phi^{\text{max}, s}$ take $s$-dimensional input.
Define a dummy network $\Phi^{\text{dummy}} \defeq \qty( \begin{pmatrix} I_s\\ O_{(2^t - s) \times s} \end{pmatrix}, 0_{2^t} )$.
Then, a $\max$ function with $s$ dimensional inputs is defined by
\begin{align*}
    \Phi^{\max, s} &\defeq \Phi^{\max, 2} \odot \langle\underbrace{\Phi^{\max, 2}, \Phi^{\max, 2}}_{2^1}\rangle \odot \dots \odot \langle\underbrace{\Phi^{\max, 2}, \dots, \Phi^{\max, 2}}_{2^t}\rangle \odot \Phi^{\text{dummy}}.
\end{align*}
Then, from Remark \ref{rem:concatenation} and Remark \ref{rem:parallelization_for_different_input}, we can verify the following relations:
\begin{enumerate}
    \item $W(\Phi^{\max, s}) \leq 2(2^0 \times 5 + 2^1 \times 5 + \dots + 2^t \times 5) + 2s = 42s$,
    \item $L(\Phi^{\max, s}) = 2(t + 1) + 1 < 2\log_2 s + 3$,
    \item $B(\Phi^{\max, s}) = 1$.
\end{enumerate}

\subsubsection{Approximation for Smooth Functions by Neural Networks}

We here investigate an approximation property of neural networks for several types of functions.
Namely, we provide several lemmas for the following functions: (i) general smooth functions, (ii) smooth functions on hypercubes, and (iii) smooth functions with finite layers.

\textbf{(i) Approximation for Smooth Functions}:
We discuss several approximation for $f_0 \in \mH(\beta, [0, 1]^D, M)$ by DNNs.
We review several existing lemmas, and also provide a novel approximation result.
To begin with, we cite the following lemma for convenience.
\begin{lemma}[Lemma A.8 in \citet{petersen2018optimal}]\label{lem:taylor_approx}
  Fix any $f \in \mH(\beta, D, M)$ and $\bar{x} \in [0, 1]^D$.
  Let $\bar{f}(x)$ be the Taylor polynomial of degree $\floor{\beta}$ of $f$ around $\bar{x}$, namely,
  \begin{equation*}
    \bar{f}(x) \defeq \sum_{\abs{\alpha} \leq \floor{\beta}} \frac{\partial^\alpha f(\bar{x})}{\alpha!} (x - \bar{x})^\alpha.
  \end{equation*}
  Then, $\abs{f(x) - \bar{f}(x)} \leq D^\beta M \norm{x - \bar{x}}^\beta$ holds for any $x \in [0, 1]^D$.
\end{lemma}
Note that there exists some constant $\overline{C} = \overline{C}(\beta, D, M)$ such that $\sup_{\abs{\alpha} \leq \floor{\beta}} \abs{\partial^\alpha f(\bar{x})/\alpha!} \leq \overline{C}M$ for $f \in \mH(\beta, [0, 1]^D, M)$.

Also, we cite the following lemma, which describes an approximation for a multiplication function.



\begin{lemma}[Lemma A.4 in  \citet{petersen2018optimal}]\label{lem:mul}
  Fix any $b > 0$. There are constants $s^{\mul{}} = s^{\mul{}}(\beta) \in \N$, $c_1^{\mul{}} = c_1^{\mul{}}(\beta, D, b)$, $c_2^{\mul{}} = c_2^{\mul{}}(\beta, D, b)$ and $\varepsilon_0^{\mul{}}=\varepsilon_0^{\mul{}}(\beta, D, b)$ such that for any $\varepsilon \in (0, \varepsilon_0^{\mul{}})$ and $\alpha \leq \floor{\beta}$, there is a neural network $\Phi_\varepsilon^{\mul}$ with $D$-dimensional input and $1$-dimensional output satisfying the following inequalities:
  \begin{enumerate}
    \item $\sup_{x \in [0, 1]^D} \abs{R(\Phi_\varepsilon^{\mul})(x) - x^\alpha} \leq \varepsilon$,
    \item $W(\Phi_\varepsilon^{\mul}) \leq c_1^{\mul{}} \varepsilon^{-D/b}$,
    \item $L(\Phi_\varepsilon^{\mul}) \leq (1 + \ceil{\log_2 \floor{\beta}})(11 + b/D)$,
    \item $B(\Phi_\varepsilon^{\mul}) \leq c_2^{\mul} \varepsilon^{-s^{\mul{}}}$.
  \end{enumerate}
  The constant $c_1^{\mul{}}$ is upper-bounded by $384 \beta ( 36r + 83 + 6 \cdot 4^{r+2})\cdot 6^{D/b}$ for $r := 1 + \floor{(b + 1)/(2D)}$.
\end{lemma}
Using this lemma, we construct a neural network approximating multiple taylor polynomials in each output.

Based on Lemma \ref{lem:mul} and Lemma A.5 in \citet{petersen2018optimal}, we develop an approximation result for an $m$-dimensional multiple output neural network $\Phi$.
Here, we write $R(\Phi) = (R(\Phi)_1, \dots, R(\Phi)_m)$.
\begin{lemma}[Simultaneous approximation of multiple Taylor polynomials]\label{lem:pol}
  Fix any $m \in \N$. Let $\qty{c_{\lambda, \alpha}} \subset [-B, B]$ for $1 \leq \lambda \leq m$. Let $(x_\lambda)_{\lambda=1}^m \subset [0, 1]^D$.
  Then there exist constants $c_1^{\pol{}} = c_1^{\pol{}}(\beta, D, d, B)$, $c_2^{\pol{}} = c_2^{\pol{}}(\beta, D, d, B)$, $s^{\pol{}}_1 = s^{\pol{}}_1(\beta, D, d, B)$ and $\varepsilon_0^{\pol{}}=\varepsilon_0^{\pol{}}(\beta, D, d)$ such that for any $\varepsilon \in (0, \varepsilon_0^{\pol{}})$, there is a neural network $\Phi_\varepsilon^{\pol}$ which satisfies the followings:
  \begin{enumerate}
    \item $\max_{\lambda=1,...,m}\sup_{x \in [0, 1]^D} \abs{R(\Phi_\varepsilon^{\pol})_\lambda(x) - \sum_{\abs{\alpha} < \beta} c_{\lambda, \alpha} (x - x_{\lambda})^\alpha} \leq \varepsilon$,
    \item $W(\Phi_\varepsilon^{\pol}) \leq c_1^{\pol{}} (\varepsilon^{-d/\beta} + m)$,
    \item $L(\Phi_\varepsilon^{\pol}) \leq 1 + (1 + \ceil{\log_2\beta})(11 + (1+\beta)/d)$,
    \item $B(\Phi_\varepsilon^{\pol}) \leq c_2^{\pol{}} \varepsilon^{-s_1^{\pol{}}}$.
  \end{enumerate}
\end{lemma}

\noindent \textbf{Proof of Lemma \ref{lem:pol}}
Firstly, we rewrite the target polynomial $\sum_{\abs{\alpha} < \beta} c_{\lambda, \alpha} (x - x_{\lambda})^\alpha$.
By the binomial theorem \citep{folland2013real} for example, we have
\begin{equation*}
    (x - x_\lambda)^\alpha = \sum_{\gamma \leq \alpha} \binom{\alpha}{\gamma} (-x_\lambda)^{\alpha - \gamma} x^\gamma.
\end{equation*}
Then, we evaluate the polynomial as
\begin{align*}
    \sum_{\abs{\alpha} < \beta} c_{\lambda, \alpha} (x - x_\lambda)^\alpha &= \sum_{\abs{\alpha} \leq \floor{\beta}} \qty{ \sum_{\gamma \leq \alpha} c_{\lambda, \alpha} \binom{\alpha}{\gamma} (-x_\lambda)^{\alpha - \gamma} x^\gamma }\\
    &= \sum_{\abs{\gamma} \leq \floor{\beta}} \qty{ \sum_{\gamma \leq \alpha, \abs{\alpha} \leq \floor{\beta}} c_{\lambda, \alpha} \binom{\alpha}{\gamma} (-x_\lambda)^{\alpha - \gamma} } x^\gamma\\
    &\eqdef \sum_{\abs{\gamma} \leq \floor{\beta}} \tilde c_{\lambda, \gamma} x^\gamma.
\end{align*}
Note that $\abs{\tilde c_{\lambda, \gamma}} \leq c M$ for all $\gamma$ with $\abs{\gamma} \leq \floor{\beta}$ where $c = c(\beta, D)$ is a constant. In fact since $\binom{\alpha}{\gamma} \leq \alpha_1^{\gamma_1} \dots \alpha_D^{\gamma_D}$, we can bound $\abs{\tilde c_{\lambda, \gamma}}$ by
\begin{align*}
    \abs{\tilde c_{\lambda, \gamma}} &\leq \sup_{x \in [0, 1]^D} \abs{\sum_{\abs{\alpha} \leq \floor{\beta}} \sum_{\gamma \leq \alpha} c_{\lambda, \alpha} \binom{\alpha}{\gamma} (-x_\lambda)^{\alpha-\gamma}}\\
    &\leq M \sum_{\abs{\alpha} \leq \floor{\beta}} \sum_{\gamma \leq \alpha} \alpha_1^{\gamma_1} \dots \alpha_D^{\gamma_D}\\
    &\leq M \sum_{\abs{\alpha} \leq \floor{\beta}} (\alpha_1+1)\dots(\alpha_D+1) \alpha_1^{\alpha_1} \dots \alpha_D^{\alpha_D}\\
    &\leq M D^{\floor{\beta}+1} (1+\floor{\beta})^{D(1 + \floor{\beta})}.
\end{align*}

Secondly, we define an explicit neural network to approximate the polynomial.
Write $\qty{\gamma \mid \abs{\gamma} \leq \floor{\beta}} = \qty{\gamma_1, \dots, \gamma_K}$ for some $K = K(\beta)$.
Let $\Phi^{{\tlr}, \lambda}_\varepsilon \defeq ((\tilde c_{\lambda, \gamma_1}, \dots, \tilde c_{\lambda, \gamma_K}), 0)$. Define $\varepsilon_0^{\pol{}} := \varepsilon_0^{\mul{}}/(cKM)$.
The number of parameters in Lemma \ref{lem:mul} has the exponential decay with exponent $-D/b$. In order to moderate the exponent, we define the neural network $\Phi^{{\mul}, \gamma_k}_{\varepsilon/cKM}$ as the one constructed by Lemma \ref{lem:mul} with substitution $\varepsilon \leftarrow \varepsilon/(cKM)$, $b \leftarrow (1+\floor{\beta})D/d$ and $\alpha \leftarrow \gamma_k$ for $k \in [K]$.
Then, there exist constants 
$c^{\pol{}}_3 = c^{\pol{}}_3(\beta, D, d, M)$ and $c^{\pol{}}_4 = c^{\pol{}}_4(\beta, D, d, M)$ such that
\begin{enumerate}
    \item $W(\Phi^{{\mul}, \gamma_k}_{\varepsilon/cKM}) \leq c^{\pol{}}_3 \varepsilon^{-d/(1 + \floor{\beta})}$,
    \item $L(\Phi^{{\mul}, \gamma_k}_{\varepsilon/cKM}) \leq (1 + \ceil{\log_2 \beta})(11 + (1+\beta)/d)$,
    \item $B(\Phi^{{\mul}, \gamma_k}_{\varepsilon/cKM}) \leq c^{\pol{}}_4 \varepsilon^{-s^{\mul{}}_2}$,
\end{enumerate}
holds for all $k \in [K]$. Note that since $K \leq D^0 + D^1 + \dots + D^{\floor{\beta}} \leq (1 + \floor{\beta}) D^{\floor{\beta} + 1}$, it is easily shown that
\begin{align*}
    c_3^{\pol} &\leq c_1^{\mul} D^{2d} (1 + \floor{\beta})^{Dd+d/(1 + \floor{\beta})} M^{d/(1 + \floor{\beta})},\\
    c_4^{\pol} &\leq c_2^{\mul} D^{s^{\mul} d/(1 + \floor{\beta}) + s^{\mul}(1 + \floor{\beta})} (cM)^{s^{\mul}} (1 + \floor{\beta})^{s^{\mul} + D(1 + \floor{\beta})s^{\mul}}.
\end{align*}
Finally, we define a concatenated and parallelized neural networks as
\begin{align*}
    \Phi_\varepsilon^{{\pol}, 1} &\defeq [\Phi_{\varepsilon/cKM}^{{\mul}, \gamma_1}, \Phi_{\varepsilon/cKM}^{{\mul}, \gamma_2}, \dots,  \Phi_{\varepsilon/cKM}^{{\mul}, \gamma_K}],\\
    \Phi_\varepsilon^{{\pol}, 2} &\defeq [\Phi^{\tlr, 1}_\varepsilon, \Phi^{\tlr, 2}_\varepsilon, \dots , \Phi^{\tlr, m}_\varepsilon], \\
    \Phi_\varepsilon^{\pol} &\defeq \Phi_\varepsilon^{{\pol}, 2} \odot \Phi_\varepsilon^{{\pol}, 1}.
\end{align*}
Then, we can simply obtain the error bound as
\begin{align*}
    \sup_{x \in [0, 1]^D} \abs{\sum_{\abs{\gamma} \leq \floor{\beta}} \tilde c_{\lambda, \gamma} x^\gamma - R(\Phi_\varepsilon^{\pol{}})_\lambda (x)} \leq cM K \frac{\varepsilon}{cMK} = \varepsilon.
\end{align*}
About parameters of the network $\Phi_\varepsilon^{\pol{}}$, the result in Remark \ref{rem:concatenation} and \ref{rem:parallelization} shows the following inequalities:
\begin{align*}
    W(\Phi_\varepsilon^{\pol}) &\leq 2W(\Phi_\varepsilon^{{\pol}, 2}) + 2W(\Phi_\varepsilon^{{\pol}, 1})\\
    &\leq 2Km + 2(2 K c_3^{\pol{}} \varepsilon^{-d/(1 + \floor{\beta})} + 4KD(1 + \ceil{\log_2 \beta})(11 + (1+\beta)/d),\\
    L(\Phi_\varepsilon^{\pol}) &= L(\Phi_\varepsilon^{{\pol}, 2}) + L(\Phi_\varepsilon^{{\pol}, 1}) \leq 1 + (1 + \ceil{\log_2 \beta})(11 + (1+\beta)/d),\\
    B(\Phi_\varepsilon^{\pol}) &\leq \max\qty{cM, c_4^{\pol{}} \varepsilon^{-s_2^{\pol{}}}}.
\end{align*}
Then, we obtain the statement.
\qed

\textbf{(ii) Approximation for Function on a Hypercube}:
We investigate simultaneous approximation of functions on several hypercubes, namely, we approximate set of functions from a set $\qty{(f_0 + M + 1) \1_I \mid I \in \mI}$.
To make the functions positive, we define $f_1 \defeq f_0 + M + 1$.
Notice that $f_1 \in \mH(\beta, [0, 1]^D, 2M+1)$ and $1 \leq f_1(x) \leq 2M + 1$ for any $x \in [0, 1]^D$.
Let $\mI$ be a minimum $\gamma$-covering of $\supp(\mu)$.
Consequently, $\mI$ is regarded as an ordered set and is accompanied by an index set $\Lambda = [\card{\mI}]$.
We define a bijective map $\psi: \mI \to \Lambda$ which returns a corresponding index in $\Lambda$ of $I \in \mI$.
Also, let $\Xi : \mI \to 2^{\Lambda}$ be a set-function defined by $\Xi(I) = \qty{I' \in \mI \mid (I \oplus 3\gamma/2) \cap I' \neq \phi}$.

We first approximate Taylor polynomials of H\"older class functions.
For any fixed $I \in \mI$, we define its center of $I$ as $(\iota_1, \dots, \iota_D)$.
Define a neural network $\Phi_\gamma^{\cut, I} = (2M+2, 0) \odot (A^2_\ell, -D) \odot [(A_1^1, b_1^1), \dots, (A_D^1, b_D^1)]$ where $A_\ell^1, b_\ell^1, A_\ell^2$ are defined by the following parameters:
\begin{align*}
    A_\ell^1 &\defeq \begin{pmatrix}e_\ell^\top & e_\ell^\top & e_\ell^\top & e_\ell^\top \\ 0 & 0 & 0 & 0\end{pmatrix}^\top,~~~
    b_\ell^1 \defeq \begin{pmatrix}-\iota_\ell + \gamma & -\iota_\ell + \gamma/2 & -\iota_\ell - \gamma/2 & -\iota_\ell - \gamma\end{pmatrix},
\end{align*}
and
\begin{align*}
    A_\ell^2 &\defeq (\underbrace{2/\gamma, -2/\gamma, -2/\gamma, 2/\gamma, 2/\gamma, -2/\gamma, -2/\gamma, 2/\gamma, \dots, 2/\gamma, -2/\gamma, -2/\gamma, 2/\gamma}_{4D}, 1/(2M+2)).
\end{align*}
Then, a function by the neural network $ R(\Phi_\gamma^{\cut, I}) [0, 1]^D \times \R_{\geq} \to \R_{\geq}$ has the following form
\begin{align}
  R(\Phi_\gamma^{\cut, I}) = (2M+2) \rho\qty(\sum_{\ell=1}^D \1^{I, \ell}_\gamma(x_\ell) + \frac{y}{2M+2} - D),\label{eq:cut}
\end{align}
where $\1_\gamma^{I, \ell}: \R \to [0, 1]$ is the approximated indicator function with the form:
\begin{equation*}
  \1^{I, \ell}_\gamma(z) = \begin{cases}
  0 & \text{if } z \leq \iota_\ell - \gamma,\\
  \frac{z - (\iota_\ell - \gamma)}{\gamma/2} & \text{if } \iota_\ell - \gamma < z \leq \iota_\ell - \frac{\gamma}{2},\\
  1 & \text{if } \iota_\ell - \frac{\gamma}{2} < z \leq \iota_\ell + \frac{\gamma}{2},\\
  \frac{(\iota_\ell + \gamma) - z}{\gamma/2} & \text{if } \iota_\ell + \frac{\gamma}{2} < z \leq \iota_\ell + \gamma,\\
  0 & \text{if } \iota_\ell+\gamma < z.
\end{cases}
\end{equation*}
Then, we can claim that the function $R(\Phi^{\cut, I}_\gamma)(x, y)$ approximates a function $(x,y) \mapsto y\1_I(x)$.
Its properties are summarized in the following remark:
\begin{remark}\label{rem:cut}
    For any $y \in [0, 2M+2]$, $R(\Phi^{\cut, I}_\gamma)(x, y) = y$ holds for any $x \in I$. Also $R(\Phi^{\cut, I}_\gamma)(x, y) \leq y$ holds for any $x \in I \oplus \gamma/2$.
    Also,  $R(\Phi^{\cut, I}_\gamma)(x, y) = 0$ holds for $x \not\in \I \oplus \gamma/2$ and any $y$.
Furthermore, we obtain the following properties:
\begin{enumerate}
    \item $W(\Phi^{\cut, I}_\gamma) = 24D + 6$,
    \item $L(\Phi^{\cut, I}_\gamma) = 3$,
    \item $B(\Phi^{\cut, I}_\gamma) \leq \max\qty{1, 2M+2, 1/(2M+2), D, 1+\gamma, 2/\gamma}$.
\end{enumerate}
\end{remark}

Then, we define a neural network to approximate $f_I$, which is a $\varepsilon/2$-accuracy Taylor polynomial of $f_1$.
For any $I \in \mI$ and any point $x_I \in I$, take $f_I(x)$ as a Taylor polynomial function as Lemma \ref{lem:taylor_approx} with setting $\bar{x} \leftarrow x_I$ and $f \leftarrow f_1$.
For a fixed $\varepsilon \in (0, \varepsilon_0^{\pol{}}/2)$, let $\Phi^{\pol}_{\varepsilon/2}$ be a neural network constructed in Lemma \ref{lem:pol} with $\varepsilon \leftarrow \varepsilon/2$, $m \leftarrow \card{\mI}$, $(x_\lambda)_{\lambda=1}^{m} \leftarrow (x_{\psi^{-1}(\lambda)})_{\lambda=1}^{\card{\mI}}$,
$(c_{\lambda, \alpha})_{\lambda=1}^{m} \leftarrow (\partial^\alpha f(x_{\psi^{-1}(\lambda)})/\alpha!)_{\lambda=1}^{\card{\mI}}$ and $B \leftarrow \overline{C} (2M+1)$, where $\overline{C} = \overline{C}(\beta, D, 2M+1)$ appearing in Lemma \ref{lem:taylor_approx}.
Then, we obtain
\begin{align}
  \sup_{I \in \mI} \sup_{x \in [0, 1]^D} \abs{f_{I}(x) - R(\Phi^{\pol}_{\varepsilon/2})_{\psi(I)}(x) } \leq \frac{\varepsilon}{2}. \label{ineq:pol}
\end{align}

Also, we construct a neural network to aggregate the outputs of $\Phi^{\pol}_{\varepsilon/2}$.
Let us define a neural network $\Phi^{\text{filter}, i} : \R^{D + m} \to \R^{D+1}$ which picks up the first $D$ inputs and $D + i$-th input as
\begin{equation*}
    \Phi^{\text{filter}, i} \defeq \qty(\qty(
    \begin{array}{cc}
        I_D & e_i^\top\\
        O_D & e_i^\top
    \end{array}
    ), 0_{D+1}).
\end{equation*}
Then, we define
\begin{align}
    \Phi^\text{simul}_{\varepsilon/2} \defeq [\Phi^{\cut, \psi^{-1}(1)}_\gamma \odot \Phi^{\text{filter}, 1}, \dots, \Phi^{\cut,\psi^{-1}(\card{\mI})}_\gamma \odot \Phi^{\text{filter}, \card{\mI}}] \odot [\Phi^{Id}_{D, L}, \Phi^{\pol}_{\varepsilon/2}]\label{def:phi_simul}
\end{align}
where $\Phi^{Id}_{D, L}$ is the neural network version of the identity function $\R^D \to \R^D$ with the number of layers $L = L(\Phi^{\pol}_{\varepsilon/2})$.

\textbf{(iii) Approximation of Smooth Functions with Finite Layers}:
We develop a Taylor polynomial approximation for $f_0$ on hypercubes $\mI$ with finite layers.
To avoid divergence of the number of layers of DNNs, we provide novel techniques for neural networks.
We first need the following lemma.
\begin{lemma}\label{lem:covering_division}
    Let $\mI$ be a minimum $\gamma$-covering of $\supp(\mu)$. Then, there exists a disjoint partition $\qty{\mI_i}_{i=1}^{5^D}$ of $\mI$ such that $\mI = \bigcup_{i=1}^{5^D} \mI_i$ and $d(I_j, I_k) \geq \gamma$ hold for any $I_j \neq I_k \in \mI_i$ if $\card(\mI_i) \geq 2$.
\end{lemma}

\noindent \textbf{Proof of Lemma \ref{lem:covering_division}}
    We construct a partition explicitly.
    Generate a sequence $\qty(\mI_i)_{i\geq1}$ inductively by the following procedure.
    Firstly, let $\mI_i = \qty{}$ for all $i \geq 1$. Starting from $i=1$, repeat the following procedures for each $i$.
    Choose any $I \in \qty{I' \in \mI\setminus\bigcup_{\ell=1}^{i} \mI_\ell \mid \min_{I'' \in \mI_i} d(I', I'') \geq \gamma}$ and let $\mI_i \leftarrow \mI_i \cup \qty{I}$ until we fail to take $I$.

    We now prove $\mI_{5^D+1} = \emptyset$ by contradiction. Suppose $\mI_{5^D+1} \neq \emptyset$. Then also $\mI_i \neq \emptyset$ for $1 \leq i \leq 5^D$.
    Take any $I \in \mI_{5^D+1}$. By construction, we can always take $I_i \in \mI_i$ such that $d(I, I_i) < \gamma$ for any $i \in [5^D]$.
    The set $\qty{I, I_1, \dots, I_{5^D}}$ is covered by $I \oplus 2\gamma$.
    But $I \oplus 2\gamma$ can be covered by $5^D$ hypercubes with diameter $\gamma$, which contradicts the fact that $\mI$ is a minimum covering.


    Since $\mI_{5^D+1} = \emptyset$, $\mI_{i} = \emptyset$ holds for all $i > 5^D$.
\qed

\subsubsection{Proof of Approximation Error bound}

\noindent \textbf{Proof of Theorem \ref{thm:approx_mink}}
Let $\mI$ be a minimum $\gamma$-covering of $\supp(\mu)$.
By Lemma \ref{lem:covering_division}, $\mI$ can be partitioned into $\mI_1, \dots, \mI_{5^D}$ such that $\mI = \bigcup_{i=1}^{5^D} \mI_i$ and  for all $i \in [5^D]$, $d(I_j, I_k) \geq \gamma$ for any $I_j, I_k \in \mI_i$ satisfying $I_j \neq I_k$.

We define a neural network that summate the output of $\Phi^{\text{simul}}_{\varepsilon/2}$ in each partition $\mI_i$.
We provide parameters $ A^{\text{sum}}_{ij} \defeq \1(\psi^{-1}(j) \in \mI_i)$ and $A^{\text{sum}} \defeq (A^{\text{sum}}_{ij})_{i, j} \in \R^{5^D \times \card{\mI}}$.
Then, we define a neural network $\Phi^{\text{sum}} \defeq \qty(A^{\text{sum}}, 0_{5^D})$. The function by $\Phi^{\text{sum}}$ has the following representation:
\begin{equation*}
    R(\Phi^{\text{sum}})(x_1, \dots, x_{\card{\mI}}) = \qty(\sum_{I \in \mI_1} x_{\psi(I)}, \dots, \sum_{I \in \mI_{5^D}} x_{\psi(I)}).
\end{equation*}

Then, we construct a neural network $\Phi^{f_1}_\varepsilon$ to approximate $f_1 \defeq f_0 + M + 1$.
Let $\Phi^{\text{simul}}_{\varepsilon/2}$ as defined in \eqref{def:phi_simul}.
Define a neural network $\Phi^{f_1}_\varepsilon $ as $\Phi^{f_1}_\varepsilon \defeq \Phi^{\max, 5^D} \odot \Phi^{\text{sum}} \odot \Phi^{\text{simul}}_{\varepsilon/2}$. We obtain a function with a form $R(\Phi^{f_1}_\varepsilon) = \max_{i\in [5^D]} \sum_{I \in \mI_i} R(\Phi^{\text{simul}}_{\varepsilon/2})_{\psi(I)}$.

Next, we bound an approximation error of $\Phi^{f_1}_\varepsilon$. When $x \in I$ for some $I \in \mI$,
\begin{align*}
    R(\Phi^{f_1}_\varepsilon)(x) &= \max_{I' \in \Xi(I)} R(\Phi^{\text{simul}}_{\varepsilon/2})_{\psi(I')}(x) \\
    &\leq \max_{I' \in \Xi(I)} R(\Phi^{\pol, \varepsilon/2})_{\psi(I')}(x),
\end{align*}
where we used the fact that
\begin{align*}
    \max_{i \in [5^D]} \sum_{I' \in \mI_i} R(\Phi^{\text{simul}}_{\varepsilon/2})_{\psi(I')} (x) &= \max_{I' \in \Xi(I)} R(\Phi^{\text{simul}}_{\varepsilon/2})_{\psi(I')}(x).
\end{align*}
In other words, when computing $R(\Phi^{f_1}_\varepsilon)(x)$ we only have to take maximum over the outputs of $R(\Phi^{\text{simul}}_{\varepsilon/2})(x)$ related to hypercubes near $x$. This follows from the fact that $R(\Phi^{f_1}_\varepsilon)_{\psi(I')}(x) = 0$ for $I' \not \in \Xi(I)$ and $d(I', I'') > \gamma$ holds for $I' \neq I'' \in \mI_i$ for all $i$. The last inequality follows by construction of $\Phi^{\text{simul}}_{\varepsilon/2}$.
For a further parameter tuning, we set $\gamma = D^{-1}\qty(3 M)^{-1/\beta} \varepsilon^{1/\beta}$.

Given $\varepsilon \in (0, 1)$, we can ensure $0 \leq R(\Phi^{\text{simul}}_{\varepsilon/2})_{\psi(I)}(x) \leq 2M + 2$ for all $I \in \mI$ by Remark \ref{rem:cut}, since $R(\Phi^{\text{simul}}_{\varepsilon/2})_{\psi(I)}$ approximates $f_I$ which is a $\varepsilon/2$-accuracy Taylor polynomial of $f_1 \in [1, 2M+1]$.
The error is bounded as
\begin{align*}
    &\abs{R(\Phi^{f_1}_{\varepsilon})(x) - f_1(x)}\\
    &= \max\qty{\max_{I' \in \Xi(I)} R(\Phi^{\text{simul}}_{\varepsilon/2})_{\psi(I')}(x) - f_1(x), f_1(x) - \max_{I' \in \Xi(I)} R(\Phi^{\text{simul}}_{\varepsilon/2})_{\psi(I')}(x)}\\
    &\leq \max\qty{\max_{I' \in \Xi(I)} R(\Phi^{\pol, \varepsilon/2})_{\psi(I')}(x) - f_1(x), f_1(x) - R(\Phi^{\pol, \varepsilon/2})_{\psi(I)}(x)}\\
    &\leq \max_{I' \in \Xi(I)} \abs{R(\Phi^{\pol, \varepsilon/2})_{\psi(I')}(x) - f_1(x)}\\
    &\leq \max_{I' \in \Xi(I)} \abs{R(\Phi^{\pol, \varepsilon/2})_{\psi(I')}(x) - f_{I'}(x)} + \max_{I' \in \Xi(I)} \abs{f_{I'}(x) - f_1(x)}\\
    &\leq \frac{\varepsilon}{2} + D^\beta M \qty(\frac{3\gamma}{2})^\beta = \varepsilon,
\end{align*}
where the second last inequality follows from Lemma \ref{lem:taylor_approx} because $f_{I'}$ for $I' \in \Xi(I)$ is a Taylor polynomial around some $x_{I'} \in I'$ satisfying $\norm{x_{I'} - x_I} \leq 3\gamma/2$. 
The last inequality follows from \eqref{ineq:pol} and Lemma \ref{lem:taylor_approx}.

From the result for $f_1$, we provide an approximation for $f_0$.
To the end, let us define a neural network $\Phi^{\text{mod}, M}$ as $\Phi^{\text{mod}, M} \defeq \qty(-1, M) \odot \qty(-1, 2M) \odot \qty(1, -1)$.
Its realization is $R(\Phi^{\text{mod}, M})(x) = \min(\max(1, x), 2M+1) - (M + 1)$ for any $x \in \R$. By Remark \ref{rem:concatenation} and \ref{rem:parallelization}, the following properties holds:
\begin{enumerate}
    \item $W(\Phi^{\text{mod}, M}) = 12$,
    \item $L(\Phi^{\text{mod}, M}) = 3$,
    \item $B(\Phi^{\text{mod}, M}) \leq \max\qty{2M, 1}$.
\end{enumerate}
Then, we define $\Phi^{f_0}_\varepsilon \defeq \Phi^{\text{mod}, M} \odot \Phi^{f_1}_{\varepsilon}$.
Then, an approximation error by $\Phi^{f_0}_\varepsilon$ is bounded as
\begin{align*}
    &\sup_{x \in \supp(\mu)} \abs{R(\Psi^{f_0}_\varepsilon)(x) - f_0(x)} \\
    &= \sup_{x \in \supp(\mu)} \abs{\min(\max(1, R(\Phi^{f_1}_{\varepsilon})(x), 2M+1) - (f_0(x) + M + 1)}\\
    &\leq \sup_{x \in \supp(\mu)} \abs{R(\Phi^{f_1}_{\varepsilon})(x) - f_1(x)}\\
    &\leq \varepsilon.
\end{align*}
Here, note that $\card{\mI} \leq c_\mu \gamma^{-d}$.
Combined with Remark \ref{rem:concatenation} and \ref{rem:parallelization}, $\Psi^{f_0}_\varepsilon$ has the following properties:
\begin{align*}
    W(\Psi^{f_0}_\varepsilon) &\leq 2W(\Phi^{\text{mod}, M}) + 2W(\Phi^{\max, 5^D}) \\
    &\quad + 2(\card{\mI}) W(\Phi^{\cut, \psi^{-1}(1)}_\gamma \odot \Phi^{\text{filter}, 1}) + 2 W(\Phi^{Id}_{D, L}) + 2W(\Phi^{\pol}_{\varepsilon/2})\\
    &\leq 2((50D + 17) c_\mu D^d (3M)^{d/\beta} + 2D(11 + (1+\beta)/d)c_1^{\pol} (2^{d/\beta} +  C D^d (3 M)^{d/\beta}))\varepsilon^{-d/\beta}\\
    &\quad+ 2(12 + 42 \times 5^D + 2D + 2D(11 + (1+\beta)/d)(1 + \ceil{\log_2\beta})),\\
    L(\Psi^{f_0}_\varepsilon) &= L(\Phi^{\text{mod}, M}) + L(\Phi^{\max, 5^D}) + L(\Phi^{\cut, \psi^{-1}(1)}_\gamma) + L(\Phi^{\text{filter}, 1}) + L(\Phi^{\pol}_{\varepsilon/2})\\
    &\leq 11 + 2D \log_2 5 + (11 + (1+\beta)/d)(1 + \ceil{\log_2\beta}),\\
    B(\Psi^{f_0}_\varepsilon) &\leq \max\qty{1, 2M+2, 1/(2M+2), D, 1 + \gamma, 2 / \gamma, c_4^{\pol}\varepsilon^{-s_2^{\pol}}}.
\end{align*}
By adjusting several constants, we obtain the statement.
\qed

\subsection{Proof of Bound for Generalization Error}

The proof of Theorem \ref{thm:gen_mink} follows proof techniques developed by several studies \citep{suzuki2017fast,schmidt2017nonparametric,imaizumi2018deep} with some adaptation for our setting.

\noindent \textbf{Proof of Theorem \ref{thm:gen_mink}}
Without loss of generality, we can assume Theorem \ref{thm:approx_mink} holds for any approximation accuracy $\varepsilon \in (0, 1)$.
In the beginning, we apply the optimal condition of $\hat{f}$ and derive a basic inequality.
Recall that $\hat f$ is defined as $\hat f(x) = \max\{-C_B, \min\{ C_B, \tilde f(x)\}\}$, where
\begin{align}
    \tilde{f} \in \argmin_{f \in \mF(W, L, B)} \sum_{i=1}^n ( Y_i - f(X_i))^2.
\end{align}
Then, it is easily seen that $\|\hat f - f_0\|_{L^2(\mu)} \leq \|\tilde f - f_0\|_{L^2(\mu)}$. Hence we regard $\hat f$ as the unclipped estimator $\tilde f$ without loss of generality.
By definition of $\hat f$, $\|Y - \hat f\|_n^2 \leq \norm{Y - f}_n^2$ for any $f \in \mF(W,L,D)$. By substituting $Y_i = f_0(X_i) + \xi_i$, we obtain the base inequality as
\begin{align}\label{eq:bias_variance_decomposition}
  \|\hat f - f_0\|_n^2 \leq \norm{f - f_0}_n^2 + \frac{2}{n} \sum_{i = 1}^n \xi_i \qty(\hat f(X_i) - f(X_i)).
\end{align}
To bound the two terms in \eqref{eq:bias_variance_decomposition}, we provide a neural network $\Psi^{f_0}_\varepsilon$ as constructed in the proof of Theorem \ref{thm:approx_mink} for approximating $f_0$.
Specifically, we set a triple $(W, L, B)$ as in Theorem \ref{thm:approx_mink} with accuracy $\varepsilon \leftarrow n^{d/(2\beta+d)}$.
We define $f^* = R(\Psi^{f_0}_\varepsilon)$.
Note that by construction, $|\hat f| \leq C_B$ and $\abs{f^*} \leq M$.

We divide the proof into following 3 steps.
\begin{description}
  \item \textbf{Step 1.} Derive an upper bound of $\|{\hat f - f^*}\|_{L^2(\mu)}^2$ using its empirical counterpart $\|{\hat f - f^*}\|_n^2$.
  \item \textbf{Step 2.} Evaluate the variance term $(1/n)\sum_{i=1}^n \xi_i\qty(\hat f(X_i) - f(X_i))$.
  \item \textbf{Step 3.} Combine the results of step 1 and step 2.
\end{description}

\textbf{Step 1. Upper Bound of $\|{\hat f - f^*}\|_{L^2(\mu)}^2$}:
We prepare an evaluation of the entropy number bound.
Let $\mN(\varepsilon, \mF, \norm{\cdot})$ be the minimum $\varepsilon$-covering number of $\mF$ by a norm $\norm{\cdot}$.
Similar results are well-known (e.g., \citet{anthony2009neural,schmidt2017nonparametric}). However, our setting, such as a parameter bound, is slightly different from those of the studies.
Hence, we provide the following lemma and its full proof.

\begin{lemma}[Covering entropy bound for $\mF$]\label{lem:entropy}
  Let $\mF = \mF(W, L, B)$ be a space of neural networks with the number of nonzero weights, the number of layers, and the maximum absolute value of weights bounded by $W, L$ and $B$ respectively. Then,
  \begin{align*}
    \log \mN\qty(\varepsilon, \mF(W, L, B), \norm{\cdot}_{L^\infty(\mu)}) \leq W \log\qty( \frac{2L B^{L} (W+1)^{L}}{\varepsilon} ).
  \end{align*}
\end{lemma}
Before presenting proof of Lemma \ref{lem:entropy}, we need the following preliminary result, which makes it possible to regard neural networks in $\mF(W, L, B)$ share the same dimensional parameter space.
\begin{lemma} \label{lem:form_nn}
  Let $\mF(W, L, B)$ be a class of neural networks. Define
  \begin{align*}
    S_B(p, q) &\defeq \qty{(A, b) \mid A \in [-B, B]^{p \times q}, b \in [-B, B]^p},\\
    \mG(W, L, B) &\defeq S_B(1, W) \times S_B(W, W) \times \dots \times S_B(W, D).
  \end{align*}
  Then there exists a map $Q: \mF(W, L, B) \to \mG(W, L, B)$ such that
  \begin{equation*}
    R(\Phi)(x) = A^Q_L \rho_{b^Q_{L-1}} \circ \dots \circ A^Q_2 \rho_{b^Q_1}(A^Q_1 x) + b^Q_L,
  \end{equation*}
  where $((A^Q_L, b^Q_L), \dots, (A^Q_1, b^Q_1)) = Q(R(\Phi)) \in \mG(W, L, B)$.
\end{lemma}
\noindent \textbf{Proof of Lemma \ref{lem:form_nn}}
  Take any $R(\Phi) \in \mF(W, L, B)$. Write $\Phi = ((A_L, b_L), \dots, (A_1, b_1))$ and assume $A_l \in \R^{p_l \times p_{l-1}}$ and $b_l \in \R^{p_l}$. Consider $(A_{l-1}, b_{l-1})$ for $l = 2, \dots, L$.
  Since the number of nonzero parameters are bounded by $W$, the number of nonzero parameters in $A_{l-1} x + b_{l-1}$ for any $x \in \R^{p_{l-2}}$ is at most $W$. For $p_{l-1} > W$, without loss of generality, we can assume the $W$-th, $\dots$, $p_l$-th element of $A_{l-1} x + b_{l-1}$ are $0$.
  Let $A_{l-1}' \in \R^{W \times p_{l-1}}$ be the upper-left part of $A_{l-1}$ and $A_{l}' \in \R^{p_l \times W}$ be the upper-left part of $A_{l}$.
  Also let $b_{l-1}' \in \R^{p_{l-1}}$ be the first $W$ elements of $b_{l-1}$.
  Then $A_l' (A_{l-1}' x + b_{l-1}') = A_l (A_{l-1}x + b_{l-1})$.
  For $p_{l-1} < W$, we can simply extend $A_l, A_{l-1}, b_{l-1}$ to be in $\R^{p_l \times W}, \R^{W \times p_{l-2}}, \R^W$, respectively.
  Applying this procedure multiple times yields the conclusion.
\qed

\noindent \textbf{Proof of Lemma \ref{lem:entropy}}
Firstly, consider neural networks $\Phi = Q(R(\Phi)) = ((A_L, b_L), \dots, (A_1, b_1))$ and $\Phi' = Q(R(\Phi')) = ((A_L', b_L'), \dots, (A_1', b_1'))$,
such that for each $l \in [L]$, $(A_l', b_l')$ has elements at most $\varepsilon$ apart from $(A_l, b_l)$.
Let us write $(A_l, b_l) = ((a_{ij}^l)_{ij}, (b_{i}^l)_i)$ and $(A_l', b_l') = (({a'}_{ij}^l)_{ij}, ({b'}_{i}^l)_i)$, then define functions in internal layers as
\begin{align*}
  h^l(x) &\defeq (h^l_1(x), \dots, h^l_{p_l}(x))^\top \defeq A_l x + b_l,\\
  {h'}^l(x) &\defeq (h^{'l}_1(x), \dots, h^{'l}_{p_l}(x))^\top \defeq A'_l x + b'_l,
 \end{align*}
and
\begin{align*}
  g^l(x) &\defeq (g^l_1(x), \dots,g^l_{p_l}(x))^\top \defeq h^l(x) - {h'}^l(x).
\end{align*}
For any $E \geq 0$,  we can bound supremums of the functions as
\begin{align}
  \sup_{x \in [-E, E]^{p_{l-1}}} \abs{g_i^l(x)} &\leq \sum_{j=1}^{p_{l-1}} \abs{a_{ij} - a_{ij}'} \abs{x_j} + \abs{b_i - b_i'} \notag \\
  &\leq p_{l-1} \varepsilon E + \varepsilon \leq (WE + 1)\varepsilon\leq (W + 1) E\varepsilon.\label{eq:sup_g}
\end{align}
Also, we have
\begin{align}
  \sup_{x \in [-E, E]^{p_{l-1}}} \abs{h_i^l(x)} &\leq \sum_{j=1}^{p_{l-1}} \abs{a_{ij}} \abs{x_j} + \abs{b_i} \notag \\
  &\leq p_{l-1} B E + B\leq (WE + 1)B \leq (W + 1) EB.\label{eq:sup_h}
\end{align}
Since a Lipschitz constant of the ReLU actiavtion function is $1$ for each coordinate, we can apply \eqref{eq:sup_h} repeatedly for $(h_1^1, \dots, h_D^1), \dots, (h_1^L, \dots, h_{p_L}^L)$.
Then, we obtain the bound for $\sup_{x \in [0, 1]^D} \abs{R(\Phi) - R(\Phi')}$ as
\begin{align*}
  &\sup_{x \in [0, 1]^D} \abs{R(\Phi) - R(\Phi')}\\
  &= \abs{h^L\circ\rho\circ h^{L-1}\circ \rho \circ \dots \circ \rho \circ h^2 \circ \rho \circ h^1(x) - {h'}^L\circ\rho\circ {h'}^{L-1}\circ \rho \circ \dots \circ \rho \circ {h'}^2 \circ \rho \circ {h'}^1(x)}\\
  &\leq \abs{h^L\circ\rho\circ h^{L-1}\circ \rho \circ \dots \circ \rho \circ h^2 \circ \rho \circ h^1(x) - {h}^L\circ\rho\circ {h}^{L-1}\circ \rho \circ \dots \circ \rho \circ {h}^2 \circ \rho \circ {h'}^1(x)}\\
  &+ \abs{h^L\circ\rho\circ h^{L-1}\circ \rho \circ \dots \circ \rho \circ h^2 \circ \rho \circ {h'}^1(x) - {h}^L\circ\rho\circ {h}^{L-1}\circ \rho \circ \dots \circ \rho \circ {h'}^2 \circ \rho \circ {h'}^1(x)}\\
  &\vdots\\
  &+ \abs{h^L\circ\rho\circ {h'}^{L-1}\circ \rho \circ \dots \circ \rho \circ {h'}^2 \circ \rho \circ {h'}^1(x) - {h'}^L\circ\rho\circ {h'}^{L-1}\circ \rho \circ \dots \circ \rho \circ {h'}^2 \circ \rho \circ {h'}^1(x)}\\
  &\leq L(W+1)^L B^{L-1} \varepsilon.
\end{align*}
Note that $E$ in \eqref{eq:sup_g} and \eqref{eq:sup_h} is bounded by $B^\ell \leq B^L$ for any $\ell = 1,...,L$.
Then, we discretize $W$ parameters with $\varepsilon / L(W + 1)^L B^{L-1}$ grid size. Thus we obtain the covering number bound in the statement.
\qed

Next, we bound the term $\|{\hat f - f^*}\|_{L^2(\mu)}^2$.
Let us define $N = \mN(\delta, \mF(W, L, B), \norm{\cdot}_{L^\infty(\mu)})$, and also $\qty{f_1, \dots, f_N}$ be a set of centers of the minimal $\delta$-cover of $\mF(W, L, B)$ with $\norm{\cdot}_{L^\infty(\mu)}$ norm.
Without loss of generality, we can assume $\abs{f_j} \leq M$ for all $j \in [N]$.
Take any random $f_{\hat j} \in \qty{f_1, \dots, f_N}$ so that $\|{\hat f - f_{\hat j}}\|_{L^\infty(\mu)} \leq \delta$.
By the triangle inequality, we have
\begin{align*}
    \|{\hat f - f^*}\|_{L^2(\mu)}^2 &\leq 2\|{\hat f - f_{\hat j}}\|_{L^2(\mu)}^2 + 2\|{f_{\hat j} - f^*}\|_{L^2(\mu)}^2\leq 2\delta^2 + 2\|{f_{\hat j} - f^*}\|_{L^2(\mu)}^2.
\end{align*}
We bound the term $\norm{f_j - f^*}_{L^2(\mu)}^2$ uniformly for all $j \in [N]$ in order to bound the random quantity $\|{f_{\hat j} - f^*}\|_{L^2(\mu)}^2$.
Firstly, from Bernstein's inequality, for independent and identically distributed random variables $Z_i$ satisfying $\abs{Z_i} \leq c$ and $E[Z_i] = 0$, it holds that
\begin{align*}
    P(\abs{\bar{Z}} \geq u) \leq \exp\qty(-\frac{nu^2}{2\tau^2 + 2cu/3})
\end{align*}
for any $u > 0$, where $\tau^2 \defeq \Var(Z_i)$.
Substitute $u \leftarrow \max\qty{v, (1/2)\norm{f_j - f^*}_{L^2(\mu)}^2}$, $Z_i \leftarrow (f_{j}(X_i) - f^*(X_i))^2 - E[(f_{j}(X_i) - f^*(X_i))^2]$ and $c \leftarrow 8M^2$.
Notice that
\begin{align*}
    \tau^2 &= E\qty[\qty((f_{j}(X_i) - f^*(X_i))^2 - E[(f_{j}(X_i) - f^*(X_i))^2])^2]\\
    &\leq 4M^2 \norm{f_j - f^*}_{L^2(\mu)}^2 \leq 8M^2 u,
\end{align*}
holds.
Then, for fixed $j$, we bound the tail probability of $\norm{f_j - f^*}_{n}$ as
\begin{align*}
    P\qty(\norm{f_j - f^*}_{L^2(\mu)}^2 \geq \norm{f_j - f^*}_{n}^2 + u) \leq \exp\qty(-\frac{3nv}{64M^2}).
\end{align*}
By the uniform bound argument, $\norm{f_j - f^*}_{L^2(\mu)}^2 \geq \norm{f_j - f^*}_{n}^2 + u$ holds
for all $j \in [N]$ with probability at most $N\exp(-3nv/(64M^2))$.
Substitute $v \leftarrow 64M^2 (n^{d/(2\beta + d)} + \log N) / (3n)$ together with the trivial inequality $u \leq v + (1/2)\norm{f_j - f^*}_{L^2(\mu)}^2$ leads to the following inequality
\begin{align*}
    \norm{f_j - f^*}_n^2 + u &\leq \norm{f_j - f^*}_{n}^2 + \frac{64M^2 n^{-2\beta/(2\beta + d)} }{3} + \frac{64M^2 \log N}{3n} + \frac{1}{2}\norm{f_j - f^*}_{L^2(\mu)}^2.
\end{align*}
Hence, for all $j \in [N]$, the following inequality
\begin{align*}
    \norm{f_j - f^*}_{L^2(\mu)}^2 \leq 2\norm{f_j - f^*}_{n}^2 + \frac{128M^2 n^{-2\beta/(2\beta + d)} }{3} + \frac{128M^2 \log N}{3n}
\end{align*}
holds with probability at least $1 - \exp\qty(-n^{d/(2\beta + d)})$.

Back to the inequality $\|{\hat f - f^*}\|_{L^2(\mu)}^2 \leq 2\delta^2 + 2\|{f_{\hat j} - f^*}\|_{L^2(\mu)}^2$ with $\delta \leftarrow n^{-\beta/(2\beta + d)}$ and with Lemma \ref{lem:entropy}, we obtain
\begin{align}
    &\|{\hat f - f^*}\|_{L^2(\mu)}^2 \notag \\
    &\leq 2n^{-2\beta/(2\beta + d)} + 4\|{f_{\hat j} - f^*}\|_{n}^2 + \frac{2^8M^2 n^{-2\beta/(2\beta + d)}}{3} + \frac{2^8M^2 \log N}{3n} \notag \\
    &\leq \qty( 10 + \frac{2^8 M^2}{3} ) n^{-2\beta/(2\beta + d)} + 8\|{\hat f - f^*}\|_{n}^2 + \frac{2^8M^2 W}{3n} \log\qty(2 n^{\beta/(2\beta + d)} L(W + 1)^L B^L), \label{ineq:result_step1}
\end{align}
with probability at least $1 - \exp\qty(-n^{d/(2\beta + d)})$.

\textbf{Step 2. Evaluate Variance}:
Let $\mG_\delta \defeq \{g \mid g \defeq f - f', \norm{g}_{L^\infty(\mu)} \leq \delta, f, f' \in \mF\}$.
Given the observed variables $X_1, \dots, X_n$, we regard $(1/n)\sum_{i=1}^n \xi_i g(X_i)$ as a stochastic process indexed by $g \in \mG_\delta$.
By the Gaussian concentration inequality(Theorem 2.5.8 in \citep{gine2016mathematical}),
\begin{equation}
  P\qty(\sup_{g \in \mG_\delta} \abs{\frac{1}{n} \sum_{i=1}^n \xi_i g(X_i)} \geq E\qty[\sup_{g \in \mG_\delta} \abs{\frac{1}{n} \sum_{i=1}^n \xi_i g(X_i)}] + r_2) \leq \exp{-\frac{nr_2^2}{2\sigma^2 \delta^2 }}. \label{ineq:tail}
\end{equation}
Also, by the covering entropy bound in \citep{gine2016mathematical} combined with the inequality $\log \mN(\varepsilon, \mG_\delta, \norm{\cdot}_n) \leq 2\log \mN(\varepsilon, \mF, \norm{\cdot}_{L^\infty(\mu)})$, we obtain
\begin{align}
  E\qty[\sup_{g \in \mG_\delta} \abs{\frac{1}{n} \sum_{i=1}^n \xi_i g(X_i)}]&\leq \frac{4\sqrt{2}\sigma}{\sqrt{n}} \int_0^{2\delta} \sqrt{\log(2\mN(\varepsilon, \mG, \norm{\cdot}_n))} \dd{\varepsilon} \notag \\
  &\leq \frac{4\sigma\sqrt{2W} \delta}{\sqrt{n}} \log\qty(\frac{L (W + 1)^L B^{L}}{\delta} + 1).  \label{ineq:exp}
\end{align}
Finally, combining \eqref{ineq:tail} and \eqref{ineq:exp} yields that
\begin{align*}
  \sup_{g \in \mG_\delta} \abs{\frac{1}{n} \sum_{i=1}^n \xi_i g(X_i)}& \leq \frac{4\sigma\sqrt{2W} \delta}{\sqrt{n}} \log\qty(\frac{L(W + 1)^L B^{L}}{\delta} + 1) + r_2\\
  &\leq \frac{1}{128} \delta^2 + 2^{11}\sigma^2\frac{W}{n} \qty(\log\qty(\frac{L (W + 1)^L B^{L}}{\delta} + 1))^2 + r_2,
\end{align*}
with probability at least $1 - \exp(-nr_2^2/(2\sigma^2 \delta^2))$.
Here, the last inequality follows from the inequality $xy \leq (1/32)x^2 + 16y^2$.
We substitute $r_2 \leftarrow 2^{-7}\delta^2$, then we have
\begin{align}
    \sup_{g \in \mG_\delta} \abs{\frac{1}{n} \sum_{i=1}^n \xi_i g(X_i)}& \leq \frac{1}{64} \delta^2 + 2^{11}\sigma^2\frac{W}{n} \qty(\log\qty(\frac{L (W + 1)^L B^{L}}{\delta} + 1))^2, \label{ineq:var2}
\end{align}
with probability $1 - \exp(-n \delta^2/(2^{-13}\sigma^2 ))$.


We bound $\|{\hat f - f^*}\|_n^2$ with substituting $\delta \leftarrow \max\{2^{15} \sigma^2 n^{-\beta/(2\beta+d)}, 2\|{\hat f - f_0}\|_n\}$. Here, we consider the following two cases.
Firstly, suppose that  $\|{\hat f - f^*}\|_n \leq \delta$ holds.
Then, we obtain
\begin{align*}
    &\|{\hat f - f^*}\|_n^2 \\
    &\leq 2\|{\hat f - f_0}\|_n^2 + 2\norm{f^* - f_0}_n^2\\
    &\leq 4\norm{f^* - f_0}_n^2 + 4\sup_{g \in \mG_\delta} \abs{\frac{1}{n} \sum_{i=1}^n \xi_i g(X_i)}\\
    &\leq 4\norm{f^* - f_0}_n^2 + 2^{13}\sigma^2\frac{W}{n} \qty(\log\qty(\frac{L (W + 1)^L B^{L}}{\delta} + 1))^2 + \frac{\delta^2}{16}\\
    &\leq 4\norm{f^* - f_0}_n^2 + 2^{13}\sigma^2\frac{W}{n} \qty(\log\qty(L (W + 1)^L B^{L} n^{\beta/(2\beta + d)} + 1))^2 + 2^{26} \sigma^4 n^{-2\beta/(2\beta + d)}\\
    &\quad+ \frac{1}{2}\|{\hat f - f^*}\|_n^2 + \frac{1}{2}\norm{f^* - f_0}_n^2,
\end{align*}
where the second inequality we apply \eqref{eq:bias_variance_decomposition} with $f \leftarrow f^*$.
Therefore,
\begin{align}\label{eq:bound_step1}
    &\|{\hat f - f^*}\|_n^2 \notag \\
    &\leq 9\norm{f^* - f_0}_n^2 + 2^{14}\sigma^2\frac{W}{n} \qty(\log\qty(L (W + 1)^L B^{L}n^{\beta/(2\beta+d)} + 1))^2 + 2^{27}\sigma^4 n^{-2\beta/(2\beta + d)}.
\end{align}

Secondly, suppose that $\|{\hat f - f^*}\|_n \geq \delta$,  namely, $2\|{\hat f - f_0}\|_n \leq \|{\hat f - f^*}\|_n$ holds.
Then, we obtain
\begin{equation*}
    \|{\hat f - f^*}\|_n^2 \leq 2\|{\hat f - f_0}\|_n^2 + 2\|{f^* - f_0}\|_n^2 \leq \frac{1}{2}\|{\hat f - f^*}\|_n^2 + 2\|{f^* - f_0}\|_n^2.
\end{equation*}
Therefore, $\|{\hat f - f^*}\|_n^2 \leq 4\norm{f^* - f_0}_n^2$.
Hence the inequality \eqref{eq:bound_step1} holds.

\textbf{Step 3. Combine the Results}:
From the conclusion of \eqref{ineq:result_step1} in Step 1 and \eqref{eq:bound_step1} in Step 2,  we obtain
\begin{align*}
    \|{\hat f - f^*}\|_{L^2(\mu)}^2 &\leq \qty(2^{30}\sigma^4 + 10 + \frac{2^8 M^2}{3}) n^{-2\beta/(2\beta + d)} + 72\norm{f^* - f_0}_n^2\\
    &\quad+ 2^{17}\sigma^2\frac{W}{n} \qty(\log\qty(L (W + 1)^L B^{L}n^{\beta/(2\beta + d)} + 1))^2\\
    &\quad+ \frac{2^8M^2 W}{3n} \log\qty(2 n^{\beta/(2\beta + d)} L(W + 1)^L B^L),
\end{align*}
with probability at least $1 - 2\exp\qty(-n^{d/(2\beta + d)})$.

Under the choice of triples $(W, L, B)$ so that $\|{f^* - f_0}\|_{L^\infty(\mu)} \leq n^{-\beta/(2\beta + d)}$, the terms inside $\log$ are polynomial to $n$.
With the inequality $\|{\hat f - f_0}\|_{L^2}^2 \leq 2\|{\hat f - f^*}\|_{L^2(\mu)}^2 + 2\|{f^* - f_0}\|_{L^2(\mu)}^2$, we conclude that there exists a constant $C_1 = C_1(c_\mu, \beta, D, d, M, \sigma)$ such as
\begin{align*}
    \|{\hat f - f_0}\|_{L^2(\mu)}^2 \leq C_1 n^{-2\beta/(2\beta + d)} (1 + \log n)^2,
\end{align*}
with probability at least $1 - 2\exp(-n^{d/(2\beta + d)})$.
\qed

\subsection{Proof for Minimax Rate of Generalization Error}

In this proof, we obtain the statement by evaluating a covering number with intrinsic dimensionality and employing the minimax rate result by \citet{yang1999information}.
We write the packing number of class $\mF$ with norm $\norm{\cdot}$ as $\mS(\varepsilon, \mF, \norm{\cdot})$, which is the maximum size of $\varepsilon$-packing of $\mF$.

\begin{lemma}[Proposition 1 in Yang and Barron (1999)]\label{lem:yangandbarron}
  Let $\mF$ be any class of functions $f$ with $\sup_{f \in \mF} \abs{f} < \infty$. For the regression model $Y_i = f_0(X_i) + \xi_i$, assume $X$ and $\varepsilon$ are independent, where $X_i \sim \mu$ and $\xi_i \sim N(0, \sigma^2)$.
  Let $\varepsilon_n$ be the solution of $\varepsilon_n^2 = \log \mS(\varepsilon_n, \mF, \norm{\cdot}_{L^2(\mu)})/n$.
  Then, we have
  \begin{equation*}
    \inf_{\hat f} \sup_{f_0 \in \mF} \|\hat f - f_0\|_{L^2(\mu)} = \Theta(\varepsilon_n),
  \end{equation*}
  where $\hat f$ is any estimator based on $n$ independent and identically distributed observations $(X_1, Y_1), \dots, (X_n, Y_n)$.
\end{lemma}
To apply Lemma \ref{lem:yangandbarron} to $\mH(\beta, [0, 1]^D, M)$, we need to evaluate the covering entropy number of the smooth function class $\mH(\beta, [0, 1]^D, M)$.
For a tight evaluation of the covering entropy of the $\mH(\beta, [0, 1]^D, M)$, we introduce the following condition.
Roughly speaking, this condition states that some minimal $\epsilon$-cover can be grouped into moderate number of subgroups in which covers are neighbouring.
\begin{definition}[Concentration Condition]
  A set $E \subset \R^K$ satisfies the Concentration Condition, when there exist constants $C_1 > 0$ and $C_2 > 0$ such that for any $\varepsilon > 0$, some $\varepsilon$-cover $\qty{B_{\infty}^K(x_i, \varepsilon)}_{i=1}^T$ of $E$ satisfies the following properties:
  there exists a map $g: \qty{x_1, \dots, x_T} \to [U]$ for some $U \in \N$ such that for all $j \in [U]$,
  and for all $X \in 2^{g^{-1}(j)} \setminus \qty{\emptyset, g^{-1}(j)}$, some $y = y \in g^{-1}(j) \setminus X$ satisfy $\min_{x \in X} \norm{x - y}_{\infty} \leq \varepsilon$. Also $T \leq C_1 \mN(\varepsilon, E, \norm{\cdot}_{\infty})$ and $U \log(1/\varepsilon) \leq C_2 \mN(\varepsilon, E, \norm{\cdot}_{\infty})$ hold.
\end{definition}
In short, this condition requires an existence of some nearly minimal $\varepsilon$-cover of $E$ that can be grouped into properly concentrated parts.
To make clear this condition, we introduce the following lemma.
\begin{lemma}\label{lem:cocentration_of_mani}
Assume $\mM$ is a compact $d$-dimensional manifold in $[0, 1]^D$, namely,
assume $\mM = \bigcup_{k=1}^K \mM_k \subset [0, 1]^D$ for some $K \in \N$. Also assume for any $1 \leq k \leq K$, there exists an onto and continuously differentiable map $\psi_k : [0, 1]^{d_k} \to \mM_k$ each of which has the input dimension $d_k \in \N$.
Then, $\mM$ satisfies the Concentration Condition.
\end{lemma}

\noindent \textbf{Proof of Lemma \ref{lem:cocentration_of_mani}}
  If $\mM_k \subset [0, 1]^D$ satisfy the Concentration Condition, it is easily shown that $\bigcup_{k=1}^K \mM_k$ satisfy the Concentration Condition. So the problem is reduced to showing that for any $k \in [K]$, $\mM_k$ satisfies the Concentration Condition.

  Fix any $k \in [K]$. For simplicity, we omit the subscript $k$ from $\psi_k$, $d_k$ and $\mM_k$.
  Write $\psi = (\psi_{1}, \dots, \psi_{D})$.
  Define $L_{i} \defeq \max_{x \in [0,1]^D} \sqrt{\sum_{j=1}^d \abs{\partial \psi_{i}'(x)/ \partial x_j}^2}$.
  Applying the mean-value theorem to $\psi_{i}$ along with Cauchy-Schwartz inequality yields $\abs{\psi_{i}(x) - \psi_{i}(y)} \leq L_{i}\norm{x - y}_2$ for any $x, y \in [0, 1]^{d}$.
  By the Lipschitz continuity of $\psi = (\psi_{1}, \dots, \psi_{D})$, for any $z, w \in [0, 1]^{d}$, $\norm{\psi(x) - \psi(y)}_{\infty} \leq \sqrt{D} L \norm{x - y}_{\infty}$ where $L \defeq \max_i L_{i}$.

  Note that $[0, 1]^d$ satisfies Concentration Condition, since for any $\delta > 0$, the $\delta$-cover $\qty{B^D_{\infty}(x_i', \delta)}_{i=1}^T$ constructed by expanding the minimal $\delta/2$-cover $\qty{B^D_{\infty}(x_i', \delta/2)}_{i=1}^T$ always satisfy the property of the condition with $U = 1$.

  Fix any $\varepsilon > 0$. Since $[0, 1]^d$ satisfies Concentration condition, we can take an $\varepsilon/(\sqrt{D}L)$-cover $\qty{B^d_{\infty}(x_{i}, \varepsilon/(\sqrt{D}L))}_{i=1}^{T}$ of $[0, 1]^d$ so that for any $X \in 2^\qty{x_1, \dots, x_T} \setminus \qty{\emptyset, \qty{x_1, \dots, x_T}}$, there exists some $y \in \qty{x_1, \dots, x_T} \setminus X$ such that $\min_{x \in X} \norm{x - y}_{\infty} \leq \varepsilon/(\sqrt{D}L)$.
  Let $\mC \defeq \qty{B^D_{\infty} (\psi(x_{1}), \varepsilon), \dots, B^D_{\infty} (\psi(x_{T}), \varepsilon)}$.
  We first verify that $\mC$ is a $\varepsilon$-cover of $\mM$.
  Since $\psi$ is onto, for any $z \in \mM$, there exists some $x \in [0, 1]^d$ such that $z = \psi(x)$. For this $x$, some $y \in \qty{x_1, \dots, x_T}$ satisfies $\norm{x - y}_{\infty} \leq \varepsilon/(\sqrt{D}L)$. Thus, we obtain
  \begin{equation*}
      \norm{z - \psi(y)}_{\infty} = \norm{\psi(x) - \psi(y)}_{\infty}\leq \sqrt{D} L \norm{x - y}_{\infty} \leq \sqrt{D} L \frac{\varepsilon}{\sqrt{D}L} = \epsilon.
  \end{equation*}
  This verifies that $\mC$ is an $\varepsilon$-cover of $\mM$.

  Take any $X' \in 2^\qty{\psi(x_1), \dots, \psi(x_T)} \setminus \qty{\emptyset, \qty{\psi(x_1), \dots, \psi(x_T)}}$. Write $X' = \qty{\psi(x_{j_1}), \dots, \psi(x_{j_t})}$. By assumption, there exists some $y \in \qty{x_1, \dots, x_T} \setminus \psi^{-1}(X')$ such that $\min_{x \in \psi^{-1}(X')} \norm{x - y}_{\infty} \leq \varepsilon/(\sqrt{D}L)$ holds. Hence, for this $y$, the following holds:
  \begin{equation*}
    \min_{\psi(x) \in X'} \norm{\psi(x) - \psi(y)}_{\infty} \leq \sqrt{D}L \min_{x \in \psi^{-1}(X')} \norm{x - y}_{\infty} \leq \sqrt{D}L \frac{\varepsilon}{\sqrt{D}L} = \varepsilon.
  \end{equation*}
  This concludes the proof.
\qed

Theorem \ref{thm:minimax} is a direct consequence of the following lemma.
\begin{lemma}[Minimax optimal rate under Concentration Condition]\label{lem:minimax}
    Let $\mu$ be a probability measure on $[0, 1]^D$. Assume $\mN(\varepsilon, \supp(\mu), \norm{\cdot}_{\infty}) = \Theta(\varepsilon^{-d})$ for some $d > 0$. Also assume that $\supp(\mu)$ satisfy Concentration Condition.
    Then, the following holds:
    \begin{equation*}
        \inf_{\hat f} \sup_{f_0 \in \mF} \|\hat f - f_0\|_{L^2(\mu)} = \Theta(\varepsilon^{-\beta/(2\beta + d)}).
    \end{equation*}
\end{lemma}
\noindent \textbf{Proof of Lemma \ref{lem:minimax}}
This proof contains the following two steps: (i) derive a lower bound of $\mN(\varepsilon, \mH(\beta, [0, 1]^D, M), \norm{\cdot}_{L^\infty(\mu)})$, (ii) derive an upper bound of $\mN(\varepsilon, \mH(\beta, [0, 1]^D, M), \norm{\cdot}_{L^\infty(\mu)})$, then (iii) apply Lemma \ref{lem:yangandbarron}.

\textbf{Step (i): The lower bound}.
For the lower bound of $\mN(\varepsilon, \mH(\beta, [0, 1]^D, M), \norm{\cdot}_{L^\infty(\mu)})$, we basically follow \citep{wainwright2019high}. We first construct a packing $\qty{f_\gamma \mid \gamma \in \qty{-1, 1}^S}$ for some $S \in \N$.
Define
\begin{equation*}
    \phi(y) \defeq \begin{cases}
        c 2^{2\beta D} \prod_{j=1}^D (1/2 - y_j)^\beta (1/2 + y_j)^\beta & \text{~if~} y \in [-1/2, 1/2]^D,\\
        0 & \text{~if~} y \not\in [-1/2, 1/2]^D,
    \end{cases}
\end{equation*}
where $c = c(\beta, D, M)$ is chosen small enough so that $\phi \in \mH(\beta, [0, 1]^D, M)$ holds.

For any $\varepsilon > 0$, set $\delta = (\varepsilon / 2c)^{1/\beta}$. Consider $\delta/2$-packing of $\supp(\mu)$ as $\qty{x_i}_{i=1}^S \subset \supp(\mu)$.
Recall that $\mN(\delta, \supp(\mu), \norm{\cdot}_{\infty}) \leq S \leq \mN(\delta/2, \supp(\mu), \norm{\cdot}_{\infty})$.

For each $\gamma \in \qty{-1, 1}^S$, define the following
\begin{equation*}
    f_\gamma(x) = \sum_{i=1}^S \gamma_i \delta^\beta \phi\qty(\frac{x - x_i}{\delta}).
\end{equation*}
If $\gamma \neq \gamma'$, then for $x \in (x_{i1} - \delta/2, x_{i1} + \delta/2) \times \dots \times    (x_{iD} - \delta/2, x_{iD} + \delta/2)$, the following holds:
\begin{equation*}
    \abs{f_\gamma(x) - f_{\gamma'}(x)} = 2\delta^\beta \phi\qty(\frac{x - x_i}{\delta}).
\end{equation*}
Setting $x \leftarrow x_i$ yields
\begin{equation*}
    \abs{f_\gamma(x) - f_{\gamma'}(x)} = 2\delta^\beta c = \varepsilon.
\end{equation*}
Since $\qty{f_\gamma \mid \gamma \in \qty{-1, 1}^S}$ is an $\varepsilon$-packing of $\mH(\beta, [0, 1]^D, M)$, we obtain the lower bound of the covering number of $\mH(\beta, [0, 1]^D, M)$ as
\begin{align*}
    \log \mN(\varepsilon, \mH(\beta, [0, 1]^D, M), \norm{\cdot}_{L^\infty(\mu)}) &\geq \log 2^S\\
    &\geq \mN(2(\varepsilon/2c)^{1/\beta}, [0, 1]^D, \norm{\cdot}_{\infty}) \log 2\\
    &\gtrsim \varepsilon^{-d/\beta},
\end{align*}
where the last inequality follows from the assumption $\mN(\varepsilon, \supp(\mu), \norm{\cdot}_{\infty}) = \Theta(\varepsilon^{-d})$.

\textbf{Step (ii): The upper bound}.
For the upper bound of $\mN(\varepsilon, \mH(\beta, [0, 1]^D, M), \norm{\cdot}_{L^\infty(\mu)})$. We modify the Theorem 2.7.1 in \citet{van1996weak}.

For a preparation, we define several notions to form a covering set with its cardinality $\mN(\varepsilon, \mH(\beta, [0, 1]^D, M), \norm{\cdot}_{L^\infty(\mu)})$.
Take the minimal $\delta$-cover $(x_i)_{i=1}^T \subset \supp(\mu)$, where $T = \mN(\delta, \supp(\mu), \norm{\cdot}_{\infty})$.
Note that $T = \Theta(\delta^{-d})$ holds by the setting.
For a multi-index $k = (k_1, \dots, k_D)$ with $k \leq \beta$, define the operators $A_k, B_k$ as
\begin{align*}
  A_k f &\defeq (\floor{D^k f(x_1)/\delta^{\beta-\abs{k}}}, \dots, \floor{D^k f(x_T)/\delta^{\beta-\abs{k}}}),\mbox{~and~}B_k f \defeq \delta^{\beta - \abs{k}} A_k f.
\end{align*}
If $A_k f = A_k g$ for all $k$ with $\abs{k} \leq \beta$, then $\norm{f - g}_{L^\infty(\mu)} \lesssim \varepsilon$.
For each $f$, we define a matrix $Af$ to provide a covering set as follows:
\begin{equation*}
  Af \defeq \begin{pmatrix}
  A_{0, 0, \dots, 0} f&
  A_{1, 0, \dots, 0} f&
  A_{0, 1, \dots, 0} f&
  \cdots&
  A_{0, 0, \dots, \beta} f
\end{pmatrix}^\top \in \R^{r \times T},
\end{equation*}
where $r$ is a combination of the multi-index, we can bound the number of row $r$ of $Af$ as $r \leq \binom{D}{0} + \binom{D}{1} + \dots + \binom{D}{\beta} \leq (\beta + 1)^D$.
Since $\abs{D^k f(x)} \leq M$ for all $x \in [0, 1]^D$,
each element in $A_k f$ takes at most $2M / \delta^{\beta - \abs{k}} + 1 \leq 2M\delta^{-\beta} + 1$ values.

Moreover, we define a smooth approximation of $D^k f(x_i)$ and evaluate its approximation error.
Suppose $\norm{x_i - x_{i'}}_{\infty} \leq \delta$ for some $i, i'$.
Since $D^k f(x_i) = \sum_{\abs{k} + \abs{l} \leq \floor{\beta}} D^{k+l} f(x_{i'}) \frac{(x_i - x_{i'})^l}{l!} + R$ with $\abs{R} \lesssim \norm{x_i - x_{i'}}_{\infty}^{\beta - \abs{k}}$, we bound the value as
\begin{align*}
  &\abs{D^k f(x_i) - \sum_{\abs{k} + \abs{l} \leq \floor{\beta}} B_{k+l} f(x_{i'}) \frac{(x_i - x_{i'})^l}{l!}}\\
  &\lesssim \sum_{\abs{k} + \abs{l} \leq \floor{\beta}} \abs{D^{k+l} f(x_i) - B_{k+l} f(x_{i'})} \frac{(x_i - x_{i'})^l}{l!} + \delta^{\beta - \abs{k}}\\
  &\leq \sum_{\abs{k} + \abs{l} \leq \floor{\beta}} \delta^{\beta - \abs{k} - \abs{l}} \frac{\delta^{\abs{l}}}{l!} + \delta^{\beta - \abs{k}}\\
  &\lesssim \delta^{\beta - \abs{k}}.
\end{align*}
Given the $i'$-th column, $i$-th column ranges over $\Theta(\delta^{\beta - \abs{k}}/{\delta^{\beta - \abs{k}}}) = \Theta(1)$.

Now, we bound the covering number by using the notions stated above.
By assumption, $\supp(\mu)$ satisfy the Concentration Condition. Thus there exist disjoint sets $X_1, \dots, X_U$ such that $X \defeq \qty{x_1, \dots, x_T} = \bigcup_{u=1}^U X_u$ and that $X_u = \qty{x_1, \dots, x_{T_u}}$ with $\norm{x_{i+1} - x_i}_{\infty} \leq \delta$ for any $i = 1, \dots, T_u-1$.
Thus
\begin{align*}
  \card\qty{Af \mid f \in \mH(\beta, [0, 1]^D, M)} \leq (2M\delta^{-\beta} + 1)^{U(\beta + 1)^D} C^{T - U},
\end{align*}
holds for some constant $C > 0$.
Substitute $\delta \leftarrow \varepsilon^{1/\beta}$, we obtain
\begin{align*}
  \log(\card\qty{Af \mid f \in \mH(\beta, [0, 1]^D, M)}) \lesssim \max\qty{U \log(\frac{1}{\varepsilon}), T - U}.
\end{align*}
Since $U \log(1/\varepsilon) =\mO(T)$, we obtain
\begin{align*}
  \log \mN\qty(\varepsilon, \mH(\beta, [0, 1]^D, M), \norm{\cdot}_{L^\infty(\mu)}) \lesssim \varepsilon^{-d/\beta}.
\end{align*}

\textbf{Step (iii): Combine the results}.
It is ready to obtain the statement.
Since it is shown that $\log \mN\qty(\varepsilon, \mH(\beta, [0, 1]^D, M), \norm{\cdot}_{L^\infty(\mu)}) = \Theta(\varepsilon^{-d/\beta})$, applying Lemma \ref{lem:yangandbarron} together with $\mS(2\varepsilon, \mH(\beta, [0, 1]^D, M), M), \norm{\cdot}_{L^\infty(\mu)}) \leq \mN(\varepsilon, \mH(\beta, [0, 1]^D, M), M), \norm{\cdot}_{L^\infty(\mu)}) \leq \mS(\varepsilon, \mH(\beta, [0, 1]^D, M), M), \norm{\cdot}_{L^\infty(\mu)})$ yields the statement.
\qed

\noindent \textbf{Proof of Theorem \ref{thm:minimax}}
By a direct application of Lemma \ref{lem:minimax}, we obtain the statement.
\qed

\vskip 0.2in

\bibliography{20-002}

\end{document}